\RequirePackage[2020-02-02]{latexrelease} 
\documentclass[twoside,11pt]{article}

\usepackage{dnd}
\usepackage{times}

\usepackage{hyperref}
\usepackage{xcolor}
\definecolor{darkblue}{rgb}{0, 0, 0.5}
\hypersetup{colorlinks=true,citecolor=darkblue, linkcolor=darkblue, urlcolor=darkblue}

\newcommand{\CITE}[1]{\citep{#1}}    
\newcommand{\autocite}[1]{\CITE{#1}}
\newcommand{\CITEA}[1]{\citet{#1}}    

\usepackage{amsmath}
\usepackage{tabularx}
\usepackage{graphicx}
\usepackage{booktabs}
\usepackage{subcaption}
\usepackage{todonotes}
\usepackage{float}

\newenvironment{EXAMPLE}{\begin{list}{}
    {\topsep      4pt
     \itemsep     .0ex
     \labelwidth  25pt
     \leftmargin  30pt
     
     }}{\end{list}}
\newcommand{\ENEW}[1]{\stepcounter{equation}
\label{#1}\item[(\theequation)\hspace{10pt}]}

\newcounter{EEXAMPLE}

\newenvironment{AEXAMPLE}{\begin{list}{}
    {\topsep      0pt
     \partopsep   0pt
     \itemsep     .0ex
     \labelwidth  2pt
     \leftmargin  0pt
     
     \usecounter{EEXAMPLE}
     }\vskip-\lastskip}{\end{list}}
\newcommand{\ENUMA}[1]{\refstepcounter{equation}
\label{#1}\item[(\theequation)] \begin{AEXAMPLE}}
\newcommand{\ENDENUMA}{\end{AEXAMPLE}}
\newcommand{\SREF}[1]{(\ref{#1})}

\newcommand{\ACRO}[1]{\textsc{#1}}
\newcommand{\ANCORA}{\ACRO{ancora}}
\newcommand{\ARRAU}{\ACRO{arrau}}
\newcommand{\BCUBED}{\ACRO{b}$^3$}

\newcommand{\BLANC}{\ACRO{blanc}}
\newcommand{\CEAF}{\ACRO{ceaf}}
\newcommand{\CL}{\ACRO{cl}}
\newcommand{\CODICRAC}{\ACRO{codi/crac}}
\newcommand{\CONLL}{\ACRO{conll}}
\newcommand{\CRAC}{\ACRO{crac}}
\newcommand{\DRT}{\ACRO{drt}}

\newcommand{\GUM}{\ACRO{gum}}
\newcommand{\FRIENDS}{\ACRO{friends}}

\newcommand{\LEA}{\ACRO{lea}}
\newcommand{\LINGEX}[1]{\textit{#1}}
\newcommand{\MUC}{\ACRO{muc}}
\newcommand{\NEWTERM}[1]{\textbf{#1}}
\newcommand{\NLP}{\ACRO{nlp}}

\newcommand{\ONTONOTES}{\ACRO{ontonotes}}
\newcommand{\PD}{\textit{Phrase Detectives}}
\newcommand{\UA}{\ACRO{ua}}

\DeclareMathOperator*{\argmax}{argmax}

\dndheading{issue(number)}{2022}{firstpage--lastpage}{Silviu Paun, Juntao Yu, Nafise Sadat Moosavi and Massimo Poesio}{10.5087/dad.DOINUMBER}

\ShortHeadings{Scoring Coreference Chains with Split-Antecedent Anaphors}{Paun, Yu, Moosavi and Poesio}
\firstpageno{1}

\begin{document}

\title{Scoring Coreference Chains with Split-Antecedent Anaphors}

\author{\name Silviu Paun\thanks{Equal contribution. Listed by alphabetical order} \email spaun3691@gmail.com \\
       \addr School of Electronic Engineering and Computer Science\\
       Queen Mary University of London
       \AND
       \name Juntao Yu$^*$ \email j.yu@essex.ac.uk \\
       \addr School of Computer Science and Electronic Engineering\\
       University of Essex
       \AND 
       \name Nafise Sadat Moosavi  \email N.S.Moosavi@sheffield.ac.uk\\
       \addr Department of Computer Science\\
       University of Sheffield
       \AND
       \name Massimo Poesio \email m.poesio@qmul.ac.uk\\
       \addr School of Electronic Engineering and Computer Science\\
       Queen Mary University of London
       }


\maketitle

\begin{abstract}%
 Anaphoric reference is an 
 aspect of language interpretation covering a variety of types of interpretation beyond the simple case of identity reference to entities introduced via nominal expressions covered by the traditional coreference task in its most recent incarnation in  {\ONTONOTES} and similar datasets.
 One of 
 these cases 
 that go beyond simple coreference is anaphoric reference to entities that must be added to the discourse model via accommodation, 
 and in particular 
 split-antecedent references to entities constructed out of other entities, as in split-antecedent plurals and in some cases of discourse deixis.
 Although this type of anaphoric reference is now annotated in many datasets, 
 systems interpreting such references  
 cannot be evaluated using the Reference coreference scorer \cite{pradhan-etal-2014-scoring}.
As part of the work towards a new scorer for anaphoric reference able to evaluate all aspects of anaphoric interpretation  
in the coverage of
the Universal Anaphora initiative, 
we propose in this paper a solution to
the 
technical problem 
of generalizing 
existing metrics for identity anaphora so that they can also be used to score cases of split-antecedents. 
This is the first such proposal in the literature on anaphora or coreference, and has been successfully used to score both split-antecedent plural references and discourse deixis in the recent {\CODICRAC} anaphora resolution in dialogue shared tasks.
\end{abstract}

\begin{keywords}
Coreference, Evaluation, Split-Antecedent Anaphors
\end{keywords}

\section{Introduction}
\label{sec:intro}
The performance of models for single-antecedent anaphora resolution on the aspects of anaphoric interpretation annotated in the 
reference 
{\ONTONOTES} dataset \cite{pradhan2012conllst}
has greatly improved in recent years \cite{wiseman2016learning,clark2016improving,lee2017end,lee2018higher,kantor-globerson-2019-coreference,joshi2019spanbert}. 
So 
the attention of the community has started to turn to 
cases of anaphora not annotated 
or not properly tested 
in {\ONTONOTES}.
Well-known examples of this trend 
is research 
on the cases of 
anaphora  whose interpretation requires some form of commonsense knowledge tested by benchmarks for the Winograd Schema Challenge \cite{rahman&ng:EMNLP2012,
liu-ijcal-17-winograd,Sakaguchi-aaai-20-winogrande},
or the pronominal anaphors that cannot be resolved purely using gender, 
for which  benchmarks such as \ACRO{gap} have been developed \cite{webster-et-al:TACL2018}.
In addition, more research has been carried out on aspects of anaphoric interpretation that go beyond identity anaphora but are  covered by datasets such as 
{\ANCORA} for Catalan and Spanish \cite{recasens&marti:LRE10}, 
{\ARRAU}  
\cite{poesio-etal-2018-anaphora,uryupina-et-al:NLEJ}
and {\GUM} for English \cite{Zeldes2017}, 
or 
the Prague Dependency Treebank for Czech \cite{nedoluzhko:LAW13}.\footnote{See  \cite{poesio-et-al:ana-book-corpora} for a more detailed survey and \cite{corefud} for a more recent, extensive update.}
These aspects include, e.g., bridging reference \cite{clark-1975-bridging,hou-et-al:CL18,hou-2020-acl,yu-etal-2020-bridging},
discourse deixis \cite{webber:91,marasovic-etal-2017-mention,kolhatkar-et-al:CL18} 
and, finally, \NEWTERM{split-antecedent anaphoric reference}, the type of anaphoric interpretation 
on which we are focusing
in this paper.
The best known example of split-antecedent are cases of plural anaphoric references 
such as pronoun \LINGEX{They} in \SREF{ex:main} \cite{eschenbach1989remarks,kamp&reyle:93}, a   plural anaphoric reference to a set composed of two or more discourse entities (John and Mary) introduced by separate  noun phrases. 

\begin{EXAMPLE}
\ENEW{ex:main}
    [John]$_1$ met [Mary]$_2$. [He]$_1$ greeted [her]$_2$. [They]$_{1,2}$ went to the movies.
\end{EXAMPLE}
Split-antecedent plural anaphors can be found in all corpora and in all languages, so an increasing number of annotation schemes cover them, 
including 
{\ANCORA} \cite{recasens&marti:LRE10},
{\ARRAU} \cite{poesio-etal-2018-anaphora, uryupina-et-al:NLEJ},
{\GUM} \cite{Zeldes2017}, 
{\FRIENDS} \cite{zhou&choi:COLING2018},
{\PD} \cite{poesio-etal-2019-crowdsourced}, 
the Prague Dependency Treebank \cite{nedoluzhko:LAW13},
and the recently created {\CODICRAC} 2021 Shared Task corpus of anaphora resolution in dialogue \cite{codi-crac-shared-task}.\footnote{See the \ACRO{CorefUD} report prepared for the Universal Anaphora initiative for an extensive discussion of the coverage of split-antecedent and other anaphors in the Universal Anaphora corpora 
\cite{corefud}.}
Split-antecedent anaphors are not always common
but e.g., in the {\FRIENDS} corpus 9\% of the mentions have more than one antecedent \cite{zhou&choi:COLING2018}.
A number of computational models for the interpretation of split-antecedent plural anaphoric references have been proposed as well \cite{vala-etal-2016-antecedents,zhou&choi:COLING2018,yu-etal-2020-plural,yu2021together}.

But in fact, other types of anaphoric reference besides split-antecedent plural references can have multiple antecedents introduced by separate text segments in exactly  the same way. 
This can happen, for instance, with discourse deixis,  as illustrated in \SREF{ex:dd:trains}, where the antecedent for discourse-deictic demonstrative \LINGEX{that} in 4.2 is the plan formed by the actions in 1.5 and 3.1, evoked by  utterances from speaker M separated by utterances from speaker S. (This example is from the \ACRO{trains} subset of the {\ARRAU} corpus \cite{uryupina-et-al:NLEJ}.)

\begin{EXAMPLE}
\ENEW{ex:dd:trains} 
\begin{tabular}{ccl}
1.1	& M & all right system\\
1.2	&   & we've got a more complicated problem\\
1.3	&   & uh\\
1.4	&   & first thing I'd like you to do\\
1.5	&   & is send engine E2 off with a boxcar to Corning to pick up oranges\\
1.6 &   & uh as soon as possible\\
2.1	& S & okay\\
3.1	& M & and while it's there it should pick up the tanker\\
4.1	& S & okay\\
4.2	&   & and [that] can get\\
4.3 &	& we can get that done by three
\end{tabular}
\end{EXAMPLE}
Split-antecedent plural references and  
discourse deixis 
are 
important from the point of view of anaphora theory 
because
they all illustrate 
the fact that reference possibilities 
in discourse models 
are not limited to entities introduced in the discourse model via nominals--indeed, 
such cases 
were one of the reasons for, e.g., Webber's development of the idea of discourse model \cite{webber:thesis}.
Unlike simple cases of identity coreference, such cases of anaphora refer to entities that need to be \NEWTERM{added to the discourse model at the point in which the anaphoric reference is encountered} via some form of inference, or \NEWTERM{accommodation} \cite{lewis:scorekeeping}.
Assigning an antecedent to split-antecedent anaphors requires a particularly simple form of inference--creating a new plural object
out of two atomic objects--but more complex cases are known requiring additional inferences  (as in,  e.g.,  context change accomodation  \cite{webber&baldwin:ACL92}). 
These types of anaphoric reference thus test  the ability of an anaphora resolution system to create antecedents \textit{ex novo} instead of choosing them from the already introduced mentions, which is what differentiates proper discourse models from simple history lists of referents.


Assessing this ability, however, requires a scorer that can evaluate the interpretation produced by a system in cases that require accommodation. 
Simplified forms of evaluation have been developed for 
bridging reference and discourse deixis and used, e.g., in the 2018 {\CRAC} Shared Task \cite{poesio-etal-2018-anaphora}, but have not become part of a standardized scorer for anaphora, and the existing metrics for discourse deixis do not work for split-antecedent discourse deixis cases such as \SREF{ex:dd:trains}. 
As for the solutions proposed for split-antecedent plural anaphors \cite{vala-etal-2016-antecedents,zhou&choi:COLING2018,yu2021together},
they either work only for split-antecedent plurals in isolation, or require a substantial redefinition of the notion of coreference chain, or only generalize one existing metrics, 
as 
discussed in detail in Section \ref{sec:relatedwork}, even though split-antecedent plurals are cases of identity anaphora after all and should therefore be in the scope of the existing Reference Coreference Scorer \cite{pradhan-etal-2014-scoring}.

The objective of this paper is to fill this gap in the literature, as part 
of the effort to develop  the new Universal Anaphora scorer \cite{yu-et-al:UA-scorer-22},\footnote{\url{https://github.com/juntaoy/universal-anaphora-scorer}}  
an extension of the Reference Coreference scorer
\cite{pradhan-etal-2014-scoring,moosavi-strube-2016-lea,poesio-etal-2018-anaphora}
that can be used to evaluate all aspects of anaphoric interpretation covered by the Universal Anaphora (\UA) initiative, including bridging reference,  discourse deixis, and some cases of accommodation. 
We start in 
Section \ref{sec:anaphora} with a summary of the types of anaphoric interpretation the new scorer is meant to cover, and  with a brief description of the metrics we set to extend in Section \ref{sec:background}. 
Crucially, these include \emph{all} the standard metrics for coreference evaluation \cite{luo&pradhan:anaphora-book:evaluation}: mention and entity-based metrics such as 
{\BCUBED} \citep{bagga-1998-bcubed} and 
{\CEAF} \citep{luo-2005-ceaf} respectively, 
as well as link-based metrics such as 
{\MUC} \citep{vilain-etal-1995-muc}, 
{\LEA} \citep{moosavi-strube-2016-lea}, and 
{\BLANC} \citep{luo-etal-2014-blanc,recasens-11-blanc-orig}. 
Our proposed extensions of the metrics, able to evaluate both single and split-antecedent references, are presented in Section \ref{sec:metrics} and illustrated with an 
example 
in Section \ref{sec:example}. 
%
%
%
Our solution is  compared with all alternative proposals regarding the scoring of split antecedent anaphors in Section \ref{sec:relatedwork}.
For a more thorough demonstration of how the metrics can be used,  
we use
the  new {\UA} scorer
incorporating these extended metrics to score both split antecedent plural reference and discourse deixis with multiple antecedents
was  
used as the official scorer for the 2021 {\CODICRAC} Shared Task on Anaphoric Interpretation in Dialogue 
\cite{codi-crac-shared-task}.\footnote{\url{https://competitions.codalab.org/competitions/30312}}  
In this paper, 
we use this scorer to 
show how the proposed generalization, unlike the proposal by \cite{zhou&choi:COLING2018} and our own proposal in \cite{yu2021together}, 
can be used to score both single and split antecedent anaphora in exactly the same way, as well as to further illustrate and analyze 
the behavior of our generalized metrics on 
the data used in previous work on split-antecedent plurals 
(Section \ref{sec:evaluation}). 
Finally, in Section \ref{sec:scoring-accommodation} we discuss how the approach proposed here could be extended to cover other types of anaphoric interpretation requiring accommodation.

\section{Anaphora, Coreference and the Scope of a Universal Anaphora Scorer }
\label{sec:anaphora}

The ultimate objective of the Universal Anaphora scorer, 
which incorporates the proposals for split-antecedent reference in this paper, 
is to assess the interpretation of all forms of anaphoric reference, not just identity anaphora. 
We briefly summarize here the types of anaphoric reference covered by the current version of the scorer, including in particular the type of reference that is the focus of the present paper because of the lack of widely accepted evaluation methods: reference to accommodated entities, as exemplified by split-antecedent anaphora.

\subsection{(Identity) Anaphora and Coreference}

%
In much {\CL} / {\NLP}  literature 
following the first Message Understanding Conference (\MUC) shared tasks  \CITE{chinchor&sundheim:95}
a distinction is made between anaphora resolution and \NEWTERM{coreference resolution}, and the term  `anaphora' is used to indicate pronominal anaphora only, 
whereas the term  coreference is used in a more general sense.
In this article, however, the terms `anaphora' and `anaphoric reference' will be used 
in the more general sense of reference to entities in the discourse model adopted in linguistics (see, e.g., \CITE{lyons:77,kamp&reyle:93})
 psycholinguistics (see, e.g., \CITE{garnham:book})
 and the pre-{\MUC} computational work (see, e.g., \cite{webber:thesis}). 
In Discourse Representation Theory (\DRT) \CITE{kamp&reyle:93}, 
for instance, proper name \LINGEX{John} in example \SREF{ex:main} introduces a new discourse entity 
1 
in the discourse model, and all subsequent mentions of John, whether using pronouns or proper names, are considered  (identity) anaphoric references to entity 
1. 
The term coreference resolution was introduced for the Message Understanding Conference \cite{chinchor&sundheim:95} to specify a rather different task  from anaphora, covering several interrelated aspects of language interpretation of interest for information extraction including not just identity anaphora resolution but also, e.g.,  the association of properties with entities in cases such as \SREF{ex:coreference}, where the  NP \LINGEX{an NLP researcher}, normally considered predicative from a linguistic perspective, would be considered `co-referent' with \LINGEX{Mary}):

\begin{EXAMPLE}
\ENEW{ex:coreference} 
      [Mary]$_1$ is [an {\NLP} researcher]$_1$.
\end{EXAMPLE}
This term `coreference' was maintained in {\NLP} even when criticism of the earlier definition \cite{van-deemter&kibble:CL} led the field to adopt a definition of the task focusing exclusively on (a subset of) identity anaphora in the (psycho-) linguistic sense, as done, e.g., in {\ONTONOTES} \cite{pradhan2012conllst}, but we will use here the more traditional linguistic terminology.

\subsection{Identity Anaphora and Beyond}

Most modern anaphoric annotation projects cover  the basic case of identity anaphora exemplified  by
\SREF{ex:identity}, where \LINGEX{Mary} and \LINGEX{her} are mentions of the same discourse entity 1 (`belong to the same coreference chain'), and \LINGEX{a new dress} and \LINGEX{it} refer to the same discourse entity 2. 
In both cases, the discourse entity is explicitly introduced with a nominal phrase.

\begin{EXAMPLE}
\ENEW{ex:identity} 
      [Mary]$_1$ bought [a new dress]$_2$ 
      but [it]$_2$ didn't fit [her]$_1$.
\end{EXAMPLE}
%
Equally, 
the several  proposed  metrics for evaluating `coreference'  (reviewed in Section \ref{sec:background}) and the Reference Coreference Scorer developed for the {\CONLL} shared task and incorporating many of these metrics \cite{pradhan-etal-2014-scoring} all focus exclusively on evaluating identity reference / identity anaphora in this sense. 

However, many other types of anaphoric reference exist beyond basic identity anaphora, including, e.g., anaphoric reference depending on relations other than identity reference, and anaphoric reference to entities not introduced using nominals. 
These types of anaphoric reference were not covered  in {\ONTONOTES} because of complexity and cost reasons (see \cite{pradhan2012conllst} as well as the discussion in \cite{zeldes:DND22}), but are covered in many of the most recent  corpora, including
{\ANCORA} \cite{recasens&marti:LRE10}, 
{\ARRAU} \cite{poesio&artstein:LREC08,poesio-etal-2018-anaphora,uryupina-et-al:NLEJ},
{\GUM} \cite{Zeldes2017},
{\PD} \cite{poesio-etal-2019-crowdsourced},
the Prague Dependency Treebank \cite{nedoluzhko:LAW13},
the \ACRO{t{\"{u}}ba-dz} corpus \cite{versley:ROLC08}
and the recently created {\CODICRAC} corpus of anaphoric reference in dialogue \cite{codi-crac-shared-task}.
The objective of the {\UA} initiative is to define a common scheme to annotate all  types of anaphoric reference and to develop a scorer that can be used to evaluate   anaphora resolution with all these datasets. 


\paragraph{Bridging references} 

Possibly the most studied type of non-identity anaphora is \NEWTERM{bridging reference} or \NEWTERM{associative anaphora}  
\cite{clark-1975-bridging} 
as in \SREF{ex:bridging}, where bridging reference / associative anaphora \LINGEX{the roof} refers to an object which is related to / associated with, but not identical to, the \LINGEX{hall}.

\begin{EXAMPLE}
\ENEW{ex:bridging} 
There was not a moment to be lost: away went Alice like the wind, and was just in time to hear it say, as it turned a corner, 'Oh my ears and whiskers, how late it's getting!' She was close behind it when she turned the corner, but the Rabbit was no longer to be seen: she found herself in [a long, low hall, which was lit up by a row of lamps hanging from \underline{[the roof]}].
\end{EXAMPLE}
Although no Reference scorer for bridging reference existed when we started work on the Universal Anaphora scorer,
approaches for evaluating bridging reference resolution did exist, in particular 
the approach from \cite{hou-et-al:CL18} already used for the {\CODICRAC} 2018 Shared Task \cite{poesio-etal-2018-anaphora}.
This method was integrated in  the Universal Anaphora scorer \cite{codi-crac-shared-task}.





\subsection{Anaphoric References Requiring Accommodation}

Another type of anaphoric reference not covered  in older  datasets such as {\ONTONOTES} and not evaluated by the Coreference reference scorers, but annotated in more advanced datasets, is anaphoric reference to entities \emph{not} introduced via nominals.
Split-antecedent anaphora, the type of anaphoric reference whose scoring is the focus of this paper, belongs to this category. 
Split-antecedent anaphora is an example of a more general class of anaphoric references that require so-called \NEWTERM{accommodation} of a new antecedent \CITE{lewis:scorekeeping,van-der-sandt:92,beaver&zeevat:accommodation}.\footnote{Note that in fact  bridging references discussed earlier require a form of accommodation as well: in \SREF{ex:bridging}  the bridging reference \LINGEX{the roof} refers to a entity  related to the already introduced \LINGEX{the hall} but has to be added to the discourse model.}
In  this paper we propose a general method for scoring split-antecedent anaphora resolution that can be extended to cover other types of anaphoric reference involving accomodation, in particular discourse deixis.
In this Section
we briefly introduce  this type of anaphoric references. 

\paragraph{Accommodation} One of the most powerful arguments for the discourse model view of anaphora, as opposed to older history-list approaches\footnote{See e.g., \cite{poesio2016book} for a review of early approaches to anaphoric interpretation.} 
is the fact that many cases of anaphoric reference cannot be interpreted with respect to the entities already introduced in the discourse model with a nominal, but require new entities   to be added to the discourse model, or accommodated \CITE{webber:thesis,kamp&reyle:93,garnham:book}.
Accommodation as conceived by \CITEA{lewis:scorekeeping} is a general operation on the discourse model which  involves adding new content that is required to process a statement being. 
For instance, \SREF{ex:accommodation:factive} presupposes that it was raining; when processing the statement, that fact has to be added to the discourse model.

\begin{EXAMPLE}
\ENEW{ex:accommodation:factive}  Mary realized it was raining.
\end{EXAMPLE}
\CITEA{webber:thesis,kamp&reyle:93,garnham:book} discuss a number of cases of anaphoric interpretation requiring new entities to be added to the discourse model to interpret anaphoric reference.

\paragraph{Split-antecedent anaphora} 
In {\ONTONOTES}, plural reference is only marked when the antecedent is mentioned by a single noun phrase. 
However, \NEWTERM{split-antecedent plural reference} is also possible \cite{eschenbach1989remarks,kamp&reyle:93}, as in \SREF{ex:split}.  
These are also cases of  plural identity reference, 
but where the antecedents are sets whose elements are two or more entities introduced by separate  noun phrases, which have to be accommodated in (i.e., added to)  the discourse model \cite{kamp&reyle:93,beaver&zeevat:accommodation}.
Such references are  annotated in, e.g., {\ARRAU} \cite{uryupina-et-al:NLEJ}, {\GUM} \cite{Zeldes2017} and  {\PD}  \cite{poesio-etal-2019-crowdsourced}.

\begin{EXAMPLE}
\ENEW{ex:split}
    [John]$_1$ met [Mary]$_2$. [He]$_1$ greeted [her]$_2$. [They]$_{1,2}$ went to the movies.
\end{EXAMPLE}
This type of anaphoric reference is not covered either by the standard Coreference Reference Scorer \cite{pradhan-etal-2014-scoring} or by the {\CODICRAC} 2018 scorer \cite{poesio-etal-2018-anaphora}, and therefore not generally attempted by anaphoric resolvers. 
A few dedicated  evaluation methods  concerned with evaluating this type of reference 
were however proposed,  are discussed in  Section \ref{sec:relatedwork}, but these proposals
either work only for split-antecedent plurals in isolation,
or require a substantial redefinition of the notion of coreference chain, or only generalize one existing metrics.
The key contribution of this paper is a method for scoring references to accommodated objects created out of antecedents separately introduced that generalizes all existing metrics and thus can be used to score both single and split antecedent anaphoric reference in exactly the same way.

\paragraph{Discourse deixis} 
A case of anaphoric reference also requiring accommodation and which has been the focus of much more {\NLP} research is \NEWTERM{discourse deixis}, or anaphora with non-nominal antecedents 
\citep{webber:91,byron:ACL02,gundel-et-al:JOLLI03,%
artstein&poesio:BRANDIAL06,kolhatkar-et-al:CL18}.
Discourse deixis, exemplified by the  reference \LINGEX{this issue} in \SREF{ex:dd:0}, is a type of abstract anaphora \CITE{asher:book} in which the antecedent is some type of abstract entity `evoked' by  the propositional content of a previous sentence. 
The evidence on discourse deixis interpretation suggests that these antecedents are not  introduced in the discourse model immediately, but are accommodated upon encountering the anaphoric reference \cite{kolhatkar-et-al:CL18}.

\begin{EXAMPLE}
\ENEW{ex:dd:0}
The municipal council had to decide [whether to balance the budget by raising revenue or cutting spending]$_i$. The council had to come to a resolution by the end of the month. [This issue]$_i$ was dividing communities across the country. \CITE{kolhatkar-et-al:CL18}
\end{EXAMPLE}
Discourse deixis is not annotated in all corpora \cite{artstein&poesio:BRANDIAL06,kolhatkar-et-al:CL18}, and when it is, the problem is simplified in a number of ways. 
In {\ONTONOTES}, only \NEWTERM{event anaphora}, a subtype of discourse deixis, is marked, 
as exemplified by \LINGEX{that} in \SREF{ex:dd}, which refers to the event of a white rabbit with pink ears running past Alice. 
(This example is from the annotated version of \textit{Alice in Wonderland} in the {\PD} corpus \citep{poesio-etal-2019-crowdsourced}.)


\begin{EXAMPLE}
\ENEW{ex:dd} ...
when suddenly a White Rabbit with pink eyes ran close by her. 
There was nothing so VERY remarkable in \underline{[that]}; nor did Alice think it so VERY much out of the way to hear the Rabbit say to itself, 'Oh dear! Oh dear! I shall be late!' 
....

\end{EXAMPLE}
The whole range of discourse deixis--i.e., including, in addition to event anaphora, references such as \LINGEX{this issue} in \SREF{ex:dd:0}--
is annotated in more recent corpora, such as, e.g.,  {\ANCORA} and {\ARRAU}.
But even in these corpora  the task of annotating discourse deixis is simplified in a number of ways.
One such simplification is that the annotators are not asked to mark as antecedent  the accommodated abstract entity, but the list of sentences or clausal units that evoke the antecedent (e.g., the clause [whether to balance the budget by raising revenue or cutting spending] in  \SREF{ex:dd}). 



No standard metric for scoring discourse deixis resolution exists, but typically systems are scored by their ability to identify the sentence evoking the antecedent.
The \NEWTERM{success @ N} metric, introduced by \CITE{kolhatkar-et-al:EMNLP13}, considers a response as correct if the key sentence is among the top N candidate response sentences identified as `antecedent' by the system.  
This metric was used by \cite{marasovic-etal-2017-mention} and in the {\CRAC} 2018 shared task \cite{poesio-etal-2018-anaphora}.
The key point from the perspective of this paper is
that this metric cannot be used to evaluate  discourse deictic references with split antecedents, 
such as the example in \SREF{ex:dd:trains}.
In the Universal Anaphora scorer, discourse deixis is scored using the same metrics used for identity anaphora; and the method introduced in  this paper for scoring split antecedent plural references  is also used to handle split antecedent cases of discourse deixis such as the one in \SREF{ex:dd:trains}.
(In fact, whereas split antecedent plurals are relatively rare, split antecedent discourse deictic references are very common e.g., in {\ARRAU}.)

\paragraph{Context Change Accommodation} As a final example of anaphoric reference requiring accommodation we will mention the cases of \NEWTERM{context change accommodation} discussed by \citet{webber&baldwin:ACL92} such as \SREF{ex:context-change-1}, where a new entity, the dough, is obtained by mixing together flour and water.

\begin{EXAMPLE}
\ENEW{ex:context-change-1} Add  [the water]$_i$ to [the flour]$_j$ little by little.\\
Then work [the dough]$_k^{i+j}$
\end{EXAMPLE}
As far as we are aware, no evaluation method has been proposed for scoring this type of anaphoric interpretation.
Although the type of accommodation required by this type of reference does not involve combining separate entities in the discourse model in sets, a slightly modified version of the proposal in this paper could be adapted to score this type of anaphoric reference, as discussed in Section \ref{sec:scoring-accommodation}.

\section{Metrics for Scoring Identity Anaphora (Coreference)}
\label{sec:background}

Before we introduce the proposed extensions that allow to additionally evaluate split-antecedent references, we briefly describe 
in this Section the existing definition of the metrics 
standardly used to evaluate single antecedent reference.
Please consult, e.g., \citep{luo&pradhan:anaphora-book:evaluation} for more in-depth discussion.


\subsection{Notation}
\label{sec:standard-notation}

The 
standard 
coreference evaluation metrics are based on the simplification that  entities can be identified with coreference chains of mentions introduced in a text, and the assumption that 
each mention refers to 
a single entity. 
We use the following notation to indicate an entity $K_i$ which can be identified with a coreference chain of mentions:
\begin{equation*}
    K_{i}  = \{ k_{i,1}, k_{i,2}, ..., k_{i,M_i} \}
\end{equation*}
We follow the standard convention and use 
$K$ and $R$ to refer to entities from the key and from the response sets, respectively. 
The key 
entities
represent the gold standard, whereas the response entities 
are the 
entities proposed by a system to be evaluated. The metrics to be described below 
adopt 
different approaches to comparing key and response entities.

\subsection{The Standard Metrics for Coreference Resolution Evaluation}

\subsubsection{Standard {\MUC}}
\label{sec:standard-muc}

{\MUC} \citep{vilain-etal-1995-muc} is a link-based metric that evaluates  response entities on the basis of the number of links they have in common with the entities in the key. 
It is standard practice to compute this information indirectly, by counting the number of missing links, and discarding them from the maximum number of possible links, as we are about to see.\footnote{In {\MUC} the maximum number of links in an entity is the minimum number of links needed to connect its mentions.} When computing 
we have:
\begin{equation*}
    \text{Recall}_{MUC} = \frac{
    \sum_i |K_i| - |\mathcal{P}(K_i; R)| 
    }{ \sum_i |K_i| - 1 }
\end{equation*}

In the equation above $\mathcal{P}(K_i; R)$ is a function called the partition function, that returns all the partitions of key entity $K_i$ with respect to the response $R$ of a system:
 \begin{equation*}
    \mathcal{P}(K_i; R) = \big\{ K_i \cap R_j \mid 1 \leq j \leq |R| \big\} \bigcup_{k_{i,u} \in K_i \setminus R} \big\{ \{k_{i,u}\} \big\}
\end{equation*}

Notice how the number of partitions indicate the number of links found in the key entities but not in the response entities. 
To compute  precision, we simply 
swap the key and response sets: 
\begin{equation*}
    \text{Precision}_{MUC} = \frac{
    \sum_i |R_i| - |\mathcal{P}(R_i; K)| 
    }{ \sum_i |R_i| - 1 }
\end{equation*}

The {\MUC} metric reports as a final value an F1-measure, which is the harmonic mean between the precision and recall presented above:
\begin{equation*}
    F = \frac{2 \times \text{Precision} \times \text{Recall}}{\text{Precision} + \text{Recall}}
\end{equation*}

\subsubsection{Standard {\BCUBED}}
\label{sec:standard-bcubed}

The {\BCUBED} metric \citep{bagga-1998-bcubed} is a mention-based metric: the evaluation measures the number of  mentions common between the entities in the key and in the response. 
Recall is computed by calculating recall for every mention, which is:

\begin{equation*}
    r(m) = \frac{  |K_i \cap R_j |  }{ |K_i| }
\end{equation*}
And then summing all these mention recalls up.
This can be done by finding all $|K_i \cap R_j |$ mentions in the intersection of key entity $K_i$ and response entity $R_j$, summing up recall for these:
\begin{equation*}
    r(i,j) = \sum_{m\in K_i \cap R_j} r(m) = 
             |K_i \cap R_j | * 
                  \frac{  |K_i \cap R_j | }{ |K_i| } = 
             \frac{  |K_i \cap R_j | ^2 }{ |K_i| }
\end{equation*}
and then summing up across all $i,j$ pairs and averaging by the total number of mentions.
The result is:
\begin{equation*}
    \text{Recall}_{B^3} = \frac{
    \sum_{i,j} \frac{ \big( |K_i \cap R_j | \big) ^2 }{ |K_i| }
    }{ \sum_i |K_i| }  
\end{equation*}
Precision is computed in a similar way,
again by swapping the key entities and the response entities.
An F1 measure can then be computed from precision and recall in the usual way.



\subsubsection{Standard {\CEAF}}
\label{sec:standard-ceaf}

{\CEAF} \citep{luo-2005-ceaf} is an entity-level metric: it aligns one to one the entities in the key with those in the response, 
and then it computes their similarity following the same function used for the alignment step.
The metric comes in two flavors, depending on the similarity function used to align and compare the entities. 
Given a key entity $K_i$ and a response entity $R_j$, \NEWTERM{mention-based} {\CEAF} is calculated using a similarity function measuring their mention overlap:
\begin{equation*}
    \phi_M (K_i, R_j) = |K_i \cap R_j | 
\end{equation*}
Whereas in
\NEWTERM{entity-based} {\CEAF}, the similarity between the two entities, 
is 
computed 
using the \ACRO{dice} coefficient:
\begin{equation*}
    \phi_E (K_i, R_j) = \frac{ 2 \Big( |K_i \cap R_j | \Big) }{ |K_i| + |R_j| }
\end{equation*}

%
At the alignment step, the Kuhn-Munkres algorithm \cite{kuhn1955hungarian,munkres1957algorithms} is used 
to find the optimal 
one-to-one mapping between the entities in the key and the response  such that their cumulative similarity is maximal. Let $K^* \subset K$ and $R^* \subset R$ be the key and the response entities for which an alignment was established,  respectively, and let $g : K^* \to R^*$ be an alignment function storing the one-to-one mappings.
Then recall is defined as the sum of the similarities over the maximal possible similarity: 

\begin{equation*}
    \text{Recall}_{CEAF} = \frac{ \sum_i \phi(K^*_i, g(K^*_i)) }{ \sum_i \phi(K_i, K_i) }
\end{equation*}
And again, 
precision is computed by swapping the entities from the key with those from the response set,
and $F1_{CEAF}$ is computed from $\text{Recall}_{CEAF}$ and $\text{Precision}_{CEAF}$  as usual.

\subsubsection{Standard {\LEA}}
\label{sec:standard-lea}

{\LEA} \citep{moosavi-strube-2016-lea} is a link-based metric which measures the resolution score of entities while taking into consideration their importance. 
So for instance for recall we have:
\begin{equation*}
    \text{Recall}_{LEA} = \frac{
    \sum_{i} \text{\small importance}(K_i) \times \text{\small resolution-score}(K_i)
    }{ \sum_i \text{\small importance}(K_i) }  
\end{equation*}
The importance of an entity could be defined in a different ways depending on the task, but for instance, 
it could be defined as being  proportional to the entity's size so that more weight is given to 
more frequently mentioned 
entities:
\begin{equation*}
    \text{importance}(K_i) = |K_i|
\end{equation*}
The resolution score, 
for recall, 
measures the proportion of links in the key that are recovered in the response: 
\begin{equation*}
    \text{resolution-score}(K_i) = \sum_j \frac{links(K_i \cap R_j)}{links(K_i)}
\end{equation*}
In {\LEA} the number of links in an entity is counted as the total number of links that can be formed between their mentions. 
For example, for an entity $K_i$, we have:
\begin{equation*}
    links(K_i) = \binom{|K_i|}{2}
\end{equation*}
As with the other metrics seen so far, Precision  is evaluated by swapping the key with the response entities,
and F1 is computed as usual. 

\subsubsection{Standard {\BLANC}}
\label{sec:standard-blanc}

{\BLANC} \citep{recasens-11-blanc-orig,luo-etal-2014-blanc} is an adaptation  for coreference resolution of the Rand Index used in clustering. 
The Rand Index is computed assessing not only a clustering algorithm's  decisions to put two entities in the same cluster, but also its decisions to put them in different clusters.
{\BLANC} adapts this approach to the case of coreference also taking into account the imbalance between the number of  \NEWTERM{coreference} (same cluster, aka coreference chain) vs.  \NEWTERM{non-coreference}  (different cluster / coreference chain) links.
To compute {\BLANC} we need to determine the number of common coreference and non-coreference links found in the key and in the response entities.
For a key $K_i$ the coreference links are computed as follows:
\begin{align*}
    C_K(i) &= \{ (k_{i,u}, k_{i,v}) \mid k_{i,u} \in K_i, k_{i,v} \in K_i, k_{i,u} \neq k_{i,v} \} 
\end{align*}
Whereas the non-coreference links between two key entities $K_i$ and $K_j$ are computed as follows:
\begin{align*}
     N_K(i,j) &= \{ (k_{i,u}, k_{j,v}) \mid k_{i,u} \in K_i, k_{j,v} \in K_j \} 
\end{align*}
It follows that the set of all coreference and non-coreference links from the keys are:
\begin{equation*}
    C_K = \cup_i C_K(i), \quad N_K = \cup_{i \neq j} N_K(i,j)  
\end{equation*}
The coreference $C_R$ and the non-coreference $N_R$ links for the response entities are computed in a similar way.
Precision and recall values are then computed for 
both coreference and non-coreference links. 
For example, for recall, we have:
\begin{equation*}
    \text{Recall}_C = \frac{|C_K \cap C_R |}{|C_K|}, \quad \text{Recall}_N = \frac{|N_K \cap N_R |}{|N_K|}
\end{equation*}
%





\section{Generalizing Standard Evaluation to Allow for Split-Antecedent References}
\label{sec:metrics}

The 
standard metrics for coreference resolution discussed in the previous Section 
all expect mentions to 
refer to 
a single entity. 
In this Section we describe an extension of these metrics that also allows  for references to multiple entities, as in the cases 
of 
split-antecedent anaphors
illustrated in \SREF{ex:main}. 
The generalization we propose follows the 
spirit of the existing metrics. 
The existing metrics assess the proposed 
single-antecedent references in the response on the basis of whether they refer to the same entity as the key;
we propose that split-antecedent references should be evaluated in the same way, i.e., on whether the entities they refer to are the same. 
This ensures that two systems which 
propose as antecedents of a split-antecedent anaphor  different mentions, but that refer 
to the same entities, will be
considered equivalent. 
Conversely, 
two systems will be considered different 
when
a split antecedent anaphor is taken to refer to different entities by  the two systems. 


The key idea on which our generalisation is based 
is 
to compare the entities 
referred to 
by a 
split-antecedent
anaphor
\textit{using the very same metrics we set to extend}.
So, for example, when scoring a system using {\MUC}, we propose to score split-antecedent anaphora resolution
according to how well 
the component entities of an accommodated set 
in the response match 
the key (gold) entities \textit{according to the same} {\MUC} \textit{metric}. 
Similarly, 
for {\BCUBED} evaluation we  use the {\BCUBED} metric,
and so on for the other metrics. 
In this way, we can 
handle 
split-antecedent anaphora evaluation within coreference evaluation without altering the existing evaluation paradigm. 
We assess both single and split-antecedent references using the very same metrics, 
preserving 
their 
individual 
strengths and weaknesses, as evolved over years of research. 
A second important characteristic of our proposed extension
is that when no split-antecedents are present,
the scores produced by the extended metrics are identical to  the scores obtained using their
standard 
formulation.

\subsection{An Extended Notation and Terminology for 
Entities}
\label{sec:notation}



In order to handle 
split-antecedent references, we generalize the notion of entity introduced in Section \ref{sec:standard-notation} to also allow entities consisting of the merge of an object constructed from the discourse model $K^o_i$ (e.g., a set constructed from the existing entities in the case of split antecedent anaphors) and a traditional coreference chain $K^m_{i}$ of mentions of that object: 
\begin{equation*}
    K_i = K^o_i \oplus K^m_{i}
\end{equation*}
This generalization of the notion of entity is used in this paper 
to provide scores for split-antecedent references, but  could also be  used to score chains including a deictic reference to an object in the visual situation, a discourse deixis, and also potentially the cases of entities introduced in the discourse model as the result of actions studied by \citet{webber&baldwin:ACL92}. 
(See Section \ref{sec:scoring-accommodation}.)



In the case of split-antecedents anaphors,
we use the notation $K^s_i$ to refer to the set that serves as antecedent for the split-antecedent reference. As per definition, $K^s_i$ is a set composed of two or more \textit{entities}:

\begin{equation*}
    K^s_{i} = \{ K_{i,1}, K_{i,2}, ..., K_{i,S_i} \}
\end{equation*}
We  use the term \NEWTERM{accommodated set} to refer to $K^s_i$, i.e., the set of two or more entities in the discourse model which was accommodated in the context to serve as the antecedent of a split antecedent \textit{anaphor}.

The antecedent entities of
a split-antecedent pronoun, in our representation above, are \textit{atomic} entities,
--i.e., entities 
all of whose mentions refer to a single antecedent. 
It is 
however 
possible 
for a split-antecedent anaphor to refer to an entity
which in turn contains a split antecedent anaphor among its mentions.
For instance, in the following example, the split antecedent anaphor \LINGEX{they all} has as  split antecedents the entity Bill and the  set
consisting of John and Mary, which in turn was accommodated in the discourse as a consequence of the split antecedent anaphor \LINGEX{they} in the second utterance.

\begin{EXAMPLE}
\ENEW{ex:split-antecedent-recursive} 
John met Mary. \\
They went to the movies, and met Bill. \\
Afterwards, \underline{they all} went  to dinner.
\end{EXAMPLE}
In this case, we recursively replace the antecedents which are 
themselves 
accommodated sets 
(i.e.,  \{John,Mary\}) with 
their element entities,
so that the outer 
accommodated set 
(the antecedent of \LINGEX{they all}) has as its 
elements 
entities referred to using single-antecedent anaphors only
(\{John,Mary,Bill\}).
Note that this 
does not lead to any loss of generality;
we resolve the split-antecedent references to sets of atomic entities as 
those are the actual antecedents
if you unpack the recursive references. 
(See  Section \ref{sec:example} for a more detailed discussion of such cases.)



We use the following notation to indicate
the set of all 
accommodated sets, 
and of all regular mentions (i.e., the single-antecedent references), respectively: 
\begin{equation*}
    K^s = \bigcup_{i} K^s_{i}, \quad 
    K^m = \bigcup_{i} K^m_{i}
\end{equation*}
The notation above was 
introduced for the entities in the key,
but the corresponding notions will also be used 
for the entities in the response. 
Also, the formulation of the coreference metrics involves computing the cardinality of an entity. 
We generalize the notion of cardinality to complex entities 
$K^s_i \oplus K^m_{i}$ 
in the obvious way as follows:
\begin{equation*}
    | K^s_i \oplus K^m_{i}| = 1 + |K^m_{i}|\\
\end{equation*}
Finally, notice how  an entity without a split-antecedent has the same representation as seen before in Section \ref{sec:standard-notation}, where only single-antecedent references were allowed:
\begin{equation*}
    K_i = K^m_i = \{ k_{i,1}, k_{i,2}, ..., k_{i,M_i} \}
\end{equation*}

\subsection{Aligning Accommodated Sets}
\label{sec:split-alignment}



All the standard coreference evaluation metrics assume an implicit `alignment' between the mentions in the key and in  the response  (the single-antecedent references). 
We say a mention in the key and one in the response  are \NEWTERM{aligned} if they share the same boundaries. 
We need to know which mentions align so we can compute the metrics: 
depending on the metric, 
the aligned mentions or the links between these mentions are used to compare the key and response entities following that metric's strategy, as discussed 
in Section \ref{sec:background}. 
If only single-antecedent anaphors are present,
and if response entities are well-formed, i.e., if 
mentions are not repeated across entities, 
each 
mention from the key is aligned with at most one mention in the response.

However, 
only aligning mentions is not sufficient
if we also have split-antecedent anaphors. 
As 
discussed 
in Section \ref{sec:notation}, split-antecedent anaphors result in the introduction of accommodated sets 
into the entities.
Thus, 
evaluating split-antecedent anaphora interpretation requires aligning these accommodated sets.
To align the accommodated sets we align their element entities.
If we do not specify 
a one-to-one alignment, the metrics would be ill-defined, i.e., contributions from multiple partially-overlapping accommodated sets
may accumulate and inflate the scores.

As briefly mentioned in the beginning of this Section, we propose to 
align 
the accommodated sets 
using the very same metric that a system is being evaluated with, for both single and split-antecedent anaphora. 
I.e., 
we propose to align 
the accommodated sets in the key entities 
and in the response entities
for the purpose of computing metric $m$ using the $F1$ scores that the same metric $m$ returns for 
the element entities of those accommodated sets.
For example, 
when computing a {\MUC} score, 
we compute the alignment score between a key accommodated set 
$K^s_i$ included in an entity $i$ and a 
response 
accommodated set 
$R^s_j$ from an entity $j$ as follows:
\begin{equation*}
    \phi (K^s_i, R^s_j) = \text{MUC}_{\text{F1}}(K^s_i, R^s_j)
\end{equation*}
The alignment process involves finding the pairs of accommodated sets 
from the key and the response that lead to the largest cumulative $F1$ score. 
Since a brute-force approach to this problem can be computationally unfeasible, we use the Kuhn-Munkres algorithm \cite{kuhn1955hungarian,munkres1957algorithms} already used in {\CEAF}, which solves alignment in polynomial time. 
Let $K^{s'} \subset K^s$ and $R^{s'} \subset R^s$ be the subsets of the aligned key and response 
accommodated sets; 
then we shall make use of the following function (and its inverse for the reverse mappings) to access the aligned split-antecedent pairs:
\begin{equation*}
    \tau : K^{s'}  \to R^{s'}
\end{equation*}

\subsection{The  Generalized Definitions}

We can now provide generalizations of 
the existing evaluation metrics 
that can be used
to score both single and split-antecedent references. 
The generalization is uniform across all metrics and only requires  an additional $\delta$ term 
responsible 
for the comparison between accommodated sets 
or 
between links involving 
accommodated sets, 
depending on the metric. 
As noted earlier in this Section, to
compute the 
score between accommodated sets for metric $m$ we compute the 
score according to $m$ between the component entities of these accommodated sets.
The accommodated sets will receive 
scores
between 0 and 1 for how well they are resolved by a system.
If entity $i$ contains no accommodated sets 
($\delta_i$ = 0) the metrics are equivalent to their original formulation (see Section \ref{sec:background}). 
When 
accommodated sets are 
perfectly resolved  ($\delta_i$ = 1) the contribution from the accommodated sets 
equals that from regular  mentions in the entity; they are after all just another 
element of the coreference chains. 
Partially resolved accommodated sets are given  scores between 0 and 1. 

\subsubsection{Generalized {\MUC}}
As just discussed, 
the {\MUC} metric is generalized to score both single and split-antecedent references by including into the original formula an additional $\delta$ term responsible for 
scoring 
the split-antecedent references:
\begin{equation*}
    \text{Recall} = \frac{
    \sum_i |K_i| - |\mathcal{P}(K^m_i; R^m)| - \delta_i
    }{ \sum_i |K_i| - 1 }
\end{equation*}
The partition function $\mathcal{P}()$ takes as arguments the regular mentions portion of the entities, i.e., the single-antecedent references 
(cfr. Section \ref{sec:notation}),
so that part of the {\MUC} formula stays unchanged. 
What changes is 
(i) the new $\delta$ term which we are about to introduce and
(ii) that the cardinality of an entity, if it contains a split-antecedent reference, is one 
greater 
than the cardinality of the entities with single-antecedent references 
only 
(see  Section \ref{sec:standard-muc} for a direct comparison).

When computing Recall, 
the 
additional 
$\delta$ term, 
measures how well the system resolves the links in the key involving 
accommodated sets. 
If an entity $K_i$ does not contain 
an accommodated set, 
then there is no such link
for the response to recover, and therefore 
$\delta_i = 0$. 
When this is the case for all entities (i.e., when no split-antecedent anaphor is present in a document) 
then generalized {\MUC} is equivalent to standard {\MUC}. 
When a key entity does contain 
an accommodated set, 
however, 
we need to score the response for how well it recovers the link to this accommodated set.
A response link is credited if one of its nodes matches 
a split-antecedent anaphor in the key
and the other consists of the aligned 
accommodated set 
(see Section \ref{sec:split-alignment} for why alignment is necessary). 
Formally, if $\exists (R^m_j, R^s_j)$ s.t. $R^m_j \in K^m_i$ and $\tau(K^s_i) = R^s_j$ then: 
\begin{equation*}
    \delta_i = 1 - \text{MUC}_{\text{Recall}} \bigg( K^s_i, R^s_j \bigg)
\end{equation*}
To reiterate, 
notice that we are comparing the 
accommodated sets in  the key and in the response
using the very same metric ($\text{MUC}_{\text{Recall}}$) we are evaluating the system for both single and split-antecedent references. 
If the key
accommodated set 
$K^s_i$ and the aligned response 
accommodated set 
$\tau(K^s_i)$ match perfectly, i.e., 
if their component entities 
are the same, 
we have $\delta_i = 0$, and so the system is fully rewarded for correctly producing the link to a split-antecedent contained by the key. 
A partial penalty is applied for an  
accommodated set in the response 
with 
{\MUC} recall $<1$, i.e., when 
the component entities of the accommodated set
are not perfectly resolved by the system. 
If there is no link in the response that satisfies the aforementioned conditions, i.e., either its nodes do not contain a  regular mention matching with the key, or 
the aligned accommodated set, 
then a missing link penalty is applied, and we set $\delta_i = 1$.

To compute {\MUC} precision we simply  swap 
the entities in the key and the response,
as per standard practice:
\begin{equation*}
    \text{Precision} = \frac{
    \sum_i |R_i| - |\mathcal{P}(R^m_i; K^m)| - \delta_i
    }{ \sum_i |R_i| - 1 }, \quad \delta_i = 1 - \text{MUC}_{\text{Precision}} \bigg( K^s_i, R^s_j \bigg)
\end{equation*}
Note also that this time the $\delta$ term is computed 
using the precision of the metric, in accordance with the principle discussed  earlier of using the very same metric 
for both single and split-antecedent references. 
{\MUC} F1 is computed as usual.

\subsubsection{Generalized {\BCUBED}}

We generalize {\BCUBED} as we just did with the {\MUC} metric,  by adding to the standard formula (discussed in Section \ref{sec:standard-bcubed}) 
a $\delta$ term responsible for evaluating the resolution of  split-antecedent references. 
For recall, we have:
\begin{equation*}
    \text{Recall} = \frac{
    \sum_{i,j} \frac{ \big( |K^m_i \cap R^m_j | + \delta_{i,j} \big) ^2 }{ |K_i| }
    }{ \sum_i |K_i| }  
\end{equation*}
When a key $K_i$ and a response $R_j$ contain an aligned accommodated set, 
i.e., $\tau(K^s_i) = R^s_j$, we need to evaluate how well the system 
resolved the 
component entities of the accommodated set.
As 
before, 
we do so using the very same metric used for the evaluation of the single-antecedents:
\begin{equation*}
    \delta_{i,j} = {\text{B}^3}_{\text{Recall}} \bigg( K^s_i, R^s_j \bigg)
\end{equation*}
A system is given full credit ($\delta_{i,j} = 1 $) when 
the component entities in the accommodated set
in the response exactly match (recall-wise) those in the key. 
Perfectly resolved split-antecedent references have a contribution to recall equal to correctly identified single-antecedent references.
When the conditions expressed earlier are not met, i.e., when the system does not produce 
an accommodated set as the interpretation of a split-antecedent anaphor
or this 
accommodated set 
is not aligned with that in the key, 
$\delta_{i,j} = 0 $.
When 
the document 
does not contain any 
accommodated sets 
generalized {\BCUBED} is equivalent to standard {\BCUBED}.

To compute precision, we proceed in the usual way and replace the keys with the responses and vice-versa. 
The {\BCUBED} F1 value is also computed in the usual way.

\subsubsection{Generalized {\CEAF}}

As 
discussed 
in Section \ref{sec:standard-ceaf}, central to computing the {\CEAF} metric are the similarity functions used to both align and compare the key and response entities. 
To extend the metric to evaluate  split-antecedent references as well we add, 
as in the other cases, 
an additional $\delta$ term to both 
the similarity function 
used in the computation of 
mention-based {\CEAF}, 
and to that used in entity-based {\CEAF}:
\begin{equation*}
    \phi_M (K_i, R_j) = |K^m_i \cap R^m_j | + \delta^{M}_{i,j}, \quad \phi_E(K_i,R_j) = \frac{ 2 \Big( |K^m_i \cap R^m_j | + \delta^{E}_{i,j} \Big) }{ |K_i| + |R_j|}
\end{equation*}
It is only when a key $K_i$ and a response $R_j$ contain 
aligned 
accommodated sets 
--i.e., $\tau(K^s_i) = R^s_j$--that we need to evaluate how well their component entities 
were resolved. 
$\delta$ for recall is computed as follows:
\begin{equation*}
    \delta^{M}_{i,j} = \text{CEAF}^{M}_{\text{Recall}} \bigg( K^s_i, R^s_j \bigg), 
    \quad
    \delta^{E}_{i,j} = \text{CEAF}^{E}_{\text{Recall}} \bigg( K^s_i, R^s_j \bigg)
\end{equation*}
When there are no 
accommodated sets 
or they are not aligned, 
$\delta_{i,j} = 0$. 
Otherwise, $\text{Recall}_{\text{CEAF}}$ is computed 
exactly as discussed in Section \ref{sec:standard-ceaf}, 
and so for $\text{Precision}_{\text{CEAF}}$ and $\text{F1}_{\text{CEAF}}$.

\subsubsection{Generalized {\LEA}}
\label{sec:lea}

To extend {\LEA} to evaluate both single and split-antecedent references we modify both the \textit{importance} and the \textit{resolution-score} functions. Starting with the former, we define the importance function to additionally include a $\beta$ term to further reward entities which contain 
an accommodated set: 
\begin{equation*}
    \text{importance}(K_i) = \beta_i |K_i|
\end{equation*}

The resolution score function is defined as in the standard version of the metric, but computed differently to also consider  split-antecedent references:
\begin{equation*}
    \text{resolution-score}(K_i) = \sum_j \frac{links(K_i \cap R_j)}{links(K_i)}
\end{equation*}
Counting the number of links in $K_i$ is trivial:  $links(K_i) = \binom{|K_i|}{2}$. 
But special attention needs to be paid when counting the number of links between the  set of mentions in common between a key $K_i$ and a response $R_j$:
\begin{equation*}
    links(K_i \cap R_j) = \binom{|K^m_i \cap R^m_j|}{2} + \delta_{i,j} \times \Big( |K^m_i \cap R^m_j|-1 \Big) 
\end{equation*}
We distinguish two types of links between the mentions common to both a key and a response entity. 
First, we have links between regular mentions present in both entities, i.e., links between single-antecedent references.
The number of these links is expressed in the first term of the equation above. 
The second type are links involving 
an accommodated set. 
When a key $K_i$ and a response $R_j$ contain aligned accommodated sets, 
i.e., $\tau(K^s_i) = R^s_j$, we need to assess how well 
their element entities 
compare:
\begin{equation*}
    \delta_{i,j} = \text{LEA}_{\text{Recall}} \bigg(K^s_i, R^s_j \bigg)
\end{equation*}
After evaluating how well the key and the response 
accommodated sets 
compare, we use this information 
to weigh 
the number of links that can have 
an accommodated set 
as a node; this is expressed in the second term from the link counting formula presented earlier. 
When there are no 
accommodated sets in 
the entities, or when they are not aligned, we set $\delta_{i,j} = 0$.

The generalized version of the metric is equivalent to the standard version presented 
in Section \ref{sec:standard-lea} when the entities in the key and response do not contain any 
accommodated sets 
(when $\delta_{i,j} = 0$, we have $ link(K_i \cap R_j) = \binom{|K^m_i \cap R^m_j|}{2}$). 
When 
accommodated sets 
do
exist, however, and they were perfectly resolved by a system, the scorer allocates full credit to each of these links involving 
accommodated sets, 
just as it does for the links between correctly identified single-antecedent references: 
when $\delta_{i,j} = 1$, 
$ links(K_i \cap R_j) = \binom{|K^m_i \cap R^m_j| + 1}{2}$. 
For imperfectly resolved 
accommodated sets 
the credit allocated to a system lies in-between the two 
extremes.

\subsubsection{Generalized {\BLANC}}

We 
saw 
back in Section \ref{sec:standard-blanc} that to compute {\BLANC} we need to establish the \textit{coreference} and the \textit{non-coreference} links found in the key and in the response entities. 
In the standard version of the metric these links are only between regular mentions, i.e., between single-antecedent references. 
In the 
generalized 
version,
in which 
entities 
may also include 
accommodated sets, 
we additionally distinguish two types of links: 
(i) links where both nodes  are 
accommodated sets,\footnote{These can occur among the non-coreference links.} 
and 
(ii) links where one node is an
accommodated set, 
and the other is a single-antecedent reference.

{\BLANC} evaluates a 
response 
by comparing the coreference and non-coreference links in the response set with those in the key. 
In standard {\BLANC} 
this is done by simply computing the intersection of these sets.
In 
the 
generalized 
version of the metric, however, we cannot do this anymore, due to the introduction of 
accommodated sets 
and the additional types of links that get created, as discussed above. 
To help us compare
different types of links we introduce a new function $\delta()$ that takes as argument two links--one from the key $(m^k_1,m^k_2)$, the other from the response $(m^r_1,m^r_2)$--and 
specifies 
how to allocate them credit. 
In short, this function will allocate full credit (i.e., a value of 1) to links 
between regular mentions that match, and partial credit (a value between 0 and 1) to those key and response links whose nodes involve aligned 
accommodated sets. 
The partial credit in this case will depend on how well 
the element entities in the accommodated sets
compare.
 The function will not allocate any credit (a value of 0) to all other pairs of links. 
We can assess how the coreference and the non-coreference links in the response compared to those 
in 
the key by evaluating the credit allocated by the function above to all pairs of links found between these sets. 
If no 
accommodated sets 
are present in the entities 
in the key and the response,
using the function as described has the same effect as the set intersection operation mentioned before used in standard {\BLANC}. 

We illustrate below how the $\delta()$ function is used to compute  recall for the non-coreference links:
\begin{equation*}
    R_N = \frac{1}{ |N_K| } \sum_{ \substack{ (m^k_1,m^k_2) \in N_K \\ (m^r_1,m^r_2) \in N_R } }  \delta \Big( (m^k_1,m^k_2), (m^r_1,m^r_2) \Big) 
\end{equation*}
A link from the key and one from the response whose nodes are regular mentions receive a credit of 1 if their mentions match. Formally, if $m^k_1, m^k_2 \in K^m, m^r_1, m^r_2 \in R^m,$ and $m^k_1 = m^r_1, m^k_2=m^r_2$ (or $m^k_1 = m^r_2, m^k_2=m^r_1$),
then:
\begin{equation*}
    \delta \Big( (m^k_1,m^k_2), (m^r_1,m^r_2) \Big) = 1
\end{equation*}
Two links 
one of whose nodes is a regular mention while the other is an accommodated set
are 
scored 
based on how well the 
accommodated sets 
compare, assuming they are aligned, and that the regular mentions match. 
In line with the rest of the metric extensions, we compare 
accommodated sets 
by comparing 
their element entities 
using the very same metric we are evaluating the system with overall, for both single and split-antecedent references. 
The current example is for the recall of the non-coreference links, so this is the metric that is used here as well. Formally, if $\exists m^k_s, m^k_m \in \{ m^k_1, m^k_2 \}$ s.t. $m^k_s \in K^s, m^k_m \in K^m$, and $ \exists m^r_s, m^r_m \in \{ m^r_1, m^r_2 \}$ s.t. $m^r_s \in R^s, m^r_m \in R^m$, and $m^k_m = m^r_m, \tau(m^k_s) = m^r_s$ then:
\begin{equation*}
    \delta \Big( (m^k_1,m^k_2), (m^r_1,m^r_2) \Big) = \text{BLANC}_{R_N}(m^k_s, m^r_s)
\end{equation*}
Two links whose nodes are aligned 
accommodated sets 
receive credit on the basis of how well 
their element entities 
compare. Formally, if $m^k_1, m^k_2 \in K^s, m^r_1, m^r_2 \in R^s,$ and $\tau(m^k_1) = m^r_1, \tau(m^k_2)=m^r_2$ \big(or $\tau(m^k_1) = m^r_2, \tau(m^k_2)=m^r_1$ \big) then:
\begin{equation*}
    \delta \Big( (m^k_1,m^k_2), (m^r_1,m^r_2) \Big) =   \text{BLANC}_{R_N} \Big(m^k_1, \tau(m^k_1) \Big) \times \text{BLANC}_{R_N} \Big(m^k_2, \tau(m^k_2) \Big)
\end{equation*}
All other links in the key and response are unrelated and receive no credit. For these, we have:
\begin{equation*}
    \delta \Big( (m^k_1,m^k_2), (m^r_1,m^r_2) \Big) = 0
\end{equation*}
All the other computations required by {\BLANC}, i.e., the precision of the non-coreference links, together with both the precision and the recall of the coreference links, are computed in a similar fashion. 
{\BLANC} then reports as its final value the arithmetic mean of the F1 values for the coreference and the non-coreference links.

\subsection{Separate Scores for the Split-antecedent References}


Split-antecedent anaphoric references 
are much rarer 
compared to single entity anaphoric references. 
As a result, they typically do not make 
a significant contribution to the overall evaluation score. 
Thus, to offer a clear picture of the performance of a system on split-antecedents, our scorer also reports separate scores 
for the split-antecedent references only.
The scores for the separate evaluation of split-antecedents are the micro-average $F_1$ of all the aligned gold and system pairs. 
Those 
accommodated sets 
for which an alignment could not be found 
(e.g., missing or spurious 
accommodated sets 
) were paired with empty sets when computing the scores.



\section{An (Artificial) Illustrative Example}
\label{sec:example}

It is not possible to evaluate the new generalized metrics by showing that they produce more intuitive outputs on some examples, as done in papers introducing  new coreference metrics, for the simple reason that no generalization of the existing metrics to cover split antecedent references was proposed so far. 
(We discuss in Section \ref{sec:relatedwork} the existing proposals regarding scoring such cases, none of which involves generalizing all existing metrics, explaining why they are not satisfactory.) 
Therefore we had to adopt a different approach to show that our generalization make sense. 

In this section, we describe in detail how the scores for 
each metric are computed under the proposed extension 
with reference to the following example, with the dual objective of illustrating how our generalization works in practice and showing that the results obtained are sensible. 
In the following Section, we compare these metrics in detail to the few existing and very partial previous proposals. 
Finally, in Section \ref{sec:evaluation},
we show that the extended metrics work in practice, in that they are effective at differentiating between systems which are intuitively better at the task and systems which are intuitively worse, 
and compare to the existing metrics, by using them to evaluate a system carrying out split-antecedent reference resolution on the same datasets used by \cite{yu2021together} and \cite{zhou&choi:COLING2018}. 


\begin{EXAMPLE}
\ENEW{ex:eval-example} 
[John]$_1$ met [Mary]$_2$ after work and asked [her]$_3$ to go see a play. \\
{[She]$_4$} liked the idea but suggested [him]$_5$ to have dinner first.\\
On [their]$_6$ way to the restaurant [they]$_7$ met [Bill]$_8$ and [Jane]$_9$.\\
{[The two]$_{10}$} were very happy to see [Bill]$_{11}$, as [they]$_{12}$ go way back. \\
{[He]$_{13}$} introduced [Jane]$_{14}$ and [all four]$_{15}$ agreed to have dinner together to catch up.
\end{EXAMPLE}
%
%
%
%
%
Only the mentions of 
entities referred to by split-antecedent anaphors
are marked in \SREF{ex:eval-example};
other mentions that would be interpreted by coreference resolvers in a normal way, 
such as a \LINGEX{a play} or \LINGEX{the restaurant},  are left out from our discussion, to keep things simpler, 
although their interpretation would also be scored  by our generalized scorer of course.

The \textit{key} entities mentioned in the example above are: 
\begin{align*}
    K_1 &= \{ \text{John}_1, \text{him}_5 \} \\
    K_2 &= \{ \text{Mary}_2, \text{her}_3, \text{She}_4 \} \\
    K_3 &=  \{K_1, K_2\} \oplus \{ \text{their}_6, \text{they}_7, \text{The two}_{10} \} \\
    K_4 &= \{ \text{Bill}_8, \text{Bill}_{11}, \text{He}_{13} \} \\
    K_5 &= \{ \text{Jane}_9, \text{Jane}_{14} \} \\
    K_6 &=  \{K_1, K_2, K_4 \} \oplus \{ \text{they}_{12} \} \\
    K_7 &=  \{K_1, K_2, K_4, K_5 \} \oplus \{ \text{all four}_{15} \}
\end{align*}
%
%
%
%
%
%
%
%
Notice first how the
accommodated set 
component of  $K_6$ is represented. 
The anaphor \LINGEX{they$_{12}$} refers to entity $K_4$ (Bill) and to entity $K_3$ 
referred to by 
\LINGEX{The two$_{10}$}, but entity $K_3$ in turn contains  
an accommodated set
with elements the entities $K_1$ (John) and $K_2$ (Mary). 
As described in Section \ref{sec:notation}, we normalize the representation of split-antecedents so that they only contain atomic entities as constituents: 
\begin{align*}
    K_6 &= \{ K_3, K_4 \} \oplus \{  \text{they}_{12} \} \\
        &=  \{ K_1, K_2, K_4 \} \oplus \{ \text{they}_{12} \} \\
\end{align*}
%
%
%
%
%
Consider next the following example response from a hypothetical coreference resolver. 
Let us call this coreference resolver `system $A$' and assume it outputs  the following response:
\begin{align*}
    R_{A,1} &= \{ \text{John}_1, \text{him}_5 \} \\
    R_{A,2} &= \{ \text{Mary}_2, \text{She}_4 \} \\
    R_{A,3} &=  \{R_{A,1}, R_{A,2} \} \oplus \{ \text{their}_6, \text{they}_7, \text{The two}_{10}, \text{they}_{12} \} \\
    R_{A,4} &= \{ \text{Bill}_8, \text{Bill}_{11} \} \\
    R_{A,5} &= \{ \text{Jane}_9, \text{Jane}_{14} \} \\
    R_{A,6} &= \{R_{A,1}, R_{A,2}, R_{A,5} \} \oplus \{  \text{all four}_{15} \}
\end{align*}
System $A$ makes several mistakes.
It
does not include mention \LINGEX{her$_{3}$} in $R_{A,2}$ 
and \LINGEX{He$_{13}$} in $R_{A,4}$  respectively, 
and mistakenly interprets \LINGEX{they$_{12}$} as a reference to John and Mary instead of to John, Mary and Bill. 
Also, System A  only produces a partially-correct interpretation of the split antecedent anaphor  
\LINGEX{all four$_{15}$} which in the key refers   to  John, Mary, Bill, and Jane, 
whereas A only recovers  3 of the 4 constituent entities. 
Using the method discussed in Section \ref{sec:split-alignment},
the 
accommodated sets in 
the response from system $A$ align with those in the key as follows:\footnote{This alignment is optimal irrespective of the similarity metric used.}
\begin{equation*}
    \tau(K^s_3) = R^s_{A,3} , \quad 
    \tau(K^s_6) = \emptyset, \quad 
    \tau(K^s_7) = R^s_{A,6},  \quad 
    \tau(R^s_{A,3}) = K^s_3,  \quad 
    \tau(R^s_{A,6}) = K^s_7 
\end{equation*}
%
%
We will compare the score assigned to system $A$ with those assigned to systems B, C and D whose output is a variation 
on how the 
accommodated set in 
$K_7$ may be resolved that raise interesting questions about the way our proposed generalizations operate:
\begin{align*}
    R_{B,6} &= \{R_{A,1}, R_{A,2}, R_{A,4} , R_{A,5} \} \oplus \{  \text{all four}_{15} \} \\
    R_{C,6} &= \{R_{A,1}, R_{A,2}, R_{A,4} \} \oplus \{  \text{all four}_{15} \} \\
    R_{D,6} &= \{R_{A,2}, R_{A,4}, R_{A,5} \} \oplus \{  \text{all four}_{15} \} \\
\end{align*}
Compared with $R_{A,6}$, 
the 
accommodated set in 
$R_{B,6}$ proposed by system B for \LINGEX{all four$_{15}$} correctly includes all 4 entities: John, Mary, Bill and Jane. 
The entity $R_{C,6}$ in C's response includes 
an accommodated set consisting of 
the  entities John, Mary and Bill, that aligns better with the 
accommodated set in
the key interpretation for \LINGEX{they$_{12}$},   $K_6$, than  with the key interpretation for \LINGEX{all four$_{15}$},  $K_7$. 
System D's interpretation for \LINGEX{all four$_{15}$},  $R_{D,6}$, also proposes 3 entities as antecedents for the anaphor, like  $R_{A,6}$, but the interpretation is  slightly worse than that proposed in A ($R_{A,1} = K_1$, but $R_{A,4}$ = $K_4 \setminus \{ \text{He}_{13} \}$).

Showing a step-by-step computation 
of
all of the metric scores, for both single and split-antecedent references, would be tedious. 
Since  the proposed generalization  does not modify the standard computation of the metrics, but just adds an additional $\delta$ term 
for the evaluation of split-antecedent references, 
we only  illustrate here  the computation of this term. 
For a step-by-step guide to the computation of the standard version of the metrics, i.e., defined only for single-antecedent references, see \citep{luo&pradhan:anaphora-book:evaluation}.

\subsection{Computing generalized MUC}


As above, we will focus on recall.
The key entities $K_1, K_2, K_4$ and $K_5$ are 
only referred to using  single-antecedent mentions. 
This means there is no link to 
an accommodated set that 
the response should recover; thus, for these entities, we have $\delta_1  = \delta_2 = \delta_4 = \delta_5 = 0$ for all  systems we are considering, A, B, C and D,
and the value of the metrics is not affected by our generalization. 
The other  entities in the key, however, do contain split-antecedent references, 
whose interpretation must be found in the responses 
and assessed.

We start with the response provided by system $A$. 
The 
accommodated set in 
entity $K_6$ cannot be optimally aligned with any accommodated set 
in $R_A$;
therefore, a missing link penalty is applied in this case, i.e., $\delta_{A,6} = 1$. 
In the case of the other two  entities in the key containing 
an accommodated set, 
$K_3$ and $K_7$,  links to 
an accommodated set 
are recovered in the response (as the conditions to have a matching regular mention and aligned 
accommodated sets 
are satisfied) and need to be evaluated. 
The 
accommodated set in 
$R_{A,3}$, 
$\{R_{A,1}, R_{A,2} \}$,
is an example of a system  identifying  the correct number of antecedents for a split antecedent anaphor, 
but resolving imperfectly 
the  antecedent entities:
\begin{align*}
    \delta_{A,3} & = 1 - \text{MUC}_{\text{Recall}} \bigg( K^s_3, R^s_{A,3} \bigg)\\
        & = 1 - \text{MUC}_{\text{Recall}} \bigg( \{ K_1, K_2 \}, \{R_{A,1}, R_{A,2} \} \bigg) \\
        & = 1 - \frac{\sum_{K_i \in \{ K_1, K_2 \}} |K_i| - |\mathcal{P}(K_i; \{ R_{A,1}, R_{A,2} \})|}{\sum_{K_i \in \{ K_1, K_2 \}} |K_i| - 1}  \\
        & = \frac{1}{3} 
\end{align*}
The entities 
included in 
the response 
accommodated set 
above have a recall of 2/3. 
The system is thus missing 1/3 to be credited a full link to the 
accommodated set in the key. 
(Note that the `mention' component of $R_{A,3}$,  $R^m_{A,3}$, is also incorrect as it includes an extra mention in comparison with $K_{3}$, but this aspect of the interpretation is not discussed here as it is not affected by our generalization.)

In the case of the 
accommodated set in 
$K_7$,
system $A$ recovers only 2 of the 3 antecedents; the missing link penalty is computed as follows:
\begin{align*}
    & \delta_{A,7} = 1 - \text{MUC}_{\text{Recall}} \bigg( K^s_7, R^s_6 \bigg) \\
       & = 1-\text{MUC}_{\text{Recall}} ( \{ K_1, K_2, K_4, K_5 \}, \{R_{A,1}, R_{A,2}, R_{A,5} \} ) \\
        & = \frac{1}{2} 
\end{align*}
%
%
Let us now compare the penalty applied to system A with  those applied to B,C and D.
The relevant $\delta$ terms are: 
\begin{equation*}
    \delta_{A,7} = \frac{1}{2}, \quad
    \delta_{B,7} = \frac{1}{3}, \quad
    \delta_{C,7} = 1, \quad
    \delta_{C,6} = 1, \quad
    \delta_{D,7} = \frac{1}{2} 
\end{equation*}
We can see that system $B$ receives a smaller penalty compared to system $A$, which makes sense considering $B$ recovers all 4 entity references. 
System $C$ is the one most heavily penalised by the scorer. First of all, 
because the optimal alignment for the 
accommodated set 
produced by $C$ is not the one in $K_7$, but  that in $K_6$,
it ends up 
completely missing the 
accommodated set 
in $K_7$ 
that the other systems get some credit for. 
Still, when computing the recall with respect to $K_6$ system $C$ continues to get a full penalty
 even though the 
 accommodated sets 
 align this time around, but a link cannot be determined because no regular mentions match. 
Finally, system $D$ gets the same penalty as system $A$. 
This is because $R_{A,1}$ and $R_{A,4}$, the entities that are different between the
accommodated sets 
from $A$ and $D$, both contribute with one link when computing {\MUC} recall. 
%
(We will see when discussing the {\BCUBED} metric below system $A$ will get 
a 
boost in score over system $D$ with this metric.)



\subsection{Computing generalized {\BCUBED}}

Again, we focus on computing 
recall,
as precision 
proceeds in the same way. 
Let us start with system $A$. 
When a key and a response entity contain aligned accommodated sets, 
we need to credit the system for how well it resolved the key 
accommodated set. 
Looking at the mappings specified by the alignment function, this only happens in two cases. 
The first case involves the 
accommodated sets 
from the $K_3$ and $R_{A,3}$ entities:
\begin{equation*}
\begin{split}
    \delta_{A,3,3} &= {\text{B}^3}_{\text{Recall}} \bigg( K^s_3, R^s_{A,3} \bigg) \\
        &= {\text{B}^3}_{\text{Recall}} \bigg( \{K_1,K_2\}, \{R_{A,1}, R_{A,2}\} \bigg) \\
        &= \frac{2}{3}
\end{split}
\end{equation*}
The second is between the 
accommodated sets in 
$K_7$ and $R_{A,6}$:
\begin{equation*}
\begin{split}
    \delta_{A,7,6} &= {\text{B}^3}_{\text{Recall}} \bigg( K^s_7, R^s_{A,6} \bigg) \\
    &= {\text{B}^3}_{\text{Recall}} \Big( \{ K_1, K_2, K_4, K_5 \}, \{R_{A,1}, R_{A,2}, R_{A,5} \} \Big) \\
    &= \frac{8}{15}
\end{split}
\end{equation*}
In all other cases, i.e., $ \forall (i, j) \in \{ 1, 2, ..., 7\} \times \{1, 2, ..., 6\} \setminus \{ (3, 3), (7, 6)\}$,
$\delta_{A,i,j} = 0$. 
We additionally compare how well system $A$ resolved the split-antecedent references with systems $B$, $C$, and $D$. The relevant $\delta$ terms are the following:
\begin{equation*}
    \delta_{A,7,6} = \frac{8}{15}, \quad
    \delta_{B,7,6} = \frac{2}{3}, \quad
    \delta_{C,7,6} = 0, \quad
    \delta_{C,6,6} = \frac{7}{12}, \quad
    \delta_{D,7,6} = \frac{7}{15} 
\end{equation*}
Among the four systems, system $B$ gets the highest credit for 
identifying all 4 entity elements of the accommodated set in 
$K_7$. System $C$ gets no credit for that 
accommodated set 
due to a different alignment with the 
accommodated set 
from entity $K_6$. Finally, system $D$ gets a slightly lower score compared with system $A$ which is intuitive since entity $R_{A,1}$ is better resolved compared with $R_{A,4}$. 

\subsection{Computing {\CEAF}}

As with {\BCUBED}, it is only when a key and a response entity 
contain aligned accommodated sets
that we need to evaluate how well the two match. 
Again, we start with the computations for system $A$. We have seen there is an alignment in two cases, one between the 
accommodated sets in 
entities $K_3$ and $R_3$:
\begin{equation*}
\begin{split}
    \delta^{M/E}_{3,3} &= \text{CEAF}^{M/E}_{\text{Recall}} \bigg( K^s_3, R^s_{A,3} \bigg) \\
        &= \begin{cases} 
            \frac{4}{5} & \text{for mention-based {\CEAF}} \\
            \frac{9}{10} & \text{for entity-based {\CEAF}}
        \end{cases}
\end{split}
\end{equation*}
The calculation is done for recall, precision is analogous. The other 
accommodated set 
alignment is between those in the $K_7$ and $R_{A,6}$ entities:
\begin{equation*}
\begin{split}
    \delta^{M/E}_{7,6} &= \text{CEAF}^{M/E}_{\text{Recall}} \bigg( K^s_7, R^s_{A,6} \bigg) \\
        &= \begin{cases} 
            \frac{3}{5} & \text{for mention-based {\CEAF}} \\
            \frac{7}{10} & \text{for entity-based {\CEAF}}
        \end{cases}
\end{split}
\end{equation*}
All other cases  either involve entities without accommodated sets, 
or 
whose accommodated sets 
do not align, so we have $\delta^{M/E}_{i,j} = 0$. 
We now introduce, for comparison, the relevant $\delta$ terms for the other systems:
\begin{align*}
    \delta^M_{A,7,6} = \frac{3}{5}, \quad
    \delta^M_{B,7,6} = \frac{4}{5}, \quad
    \delta^M_{C,7,6} = 0, \quad
    \delta^M_{C,6,6} = \frac{3}{4}, \quad
    \delta^M_{D,7,6} = \frac{3}{5} 
    \\
    \delta^E_{A,7,6} = \frac{7}{10}, \quad
    \delta^E_{B,7,6} = \frac{9}{10}, \quad
    \delta^E_{C,7,6} = 0, \quad
    \delta^E_{C,6,6} = \frac{13}{15}, \quad
    \delta^E_{D,7,6} = \frac{13}{20} 
\end{align*}
As with the other metrics presented so far system $B$ gets
the highest score
for identifying all four entities from the split-antecedent reference  in $K_7$. 
System $C$ is not allocated any credit for the accommodated set in
$K_7$, only for the one from $K_6$, because of how the split-antecedents get aligned. 
And system $D$ is found on par with system $A$ when using mention-based {\CEAF} and slightly worse (as it intuitively should) when the evaluation is conducted using entity-based {\CEAF}. This is another illustration of the strengths and weaknesses of the existing metrics for coreference evaluation which our scorer inherit. For the computation of the rest of the metrics, the observations for systems $B$, $C$, and $D$ will be similar to those expressed so far, and will be omitted. The computations will focus from now on on system $A$ exclusively just to exemplify the methodology.

\subsection{Computing {\LEA}}

When computing {\LEA}, 
for those key and response sets that contain aligned accommodated sets, 
we need to evaluate how well these 
accommodated sets 
compare. For recall, we have:
\begin{equation*}
    \delta_{i,j} = \begin{cases}
        \text{LEA}_{\text{Recall}} \Big( K^s_3, R^s_{A,3} \Big) & \text{for $i=j=3$} \\
        \text{LEA}_{\text{Recall}} \Big( K^s_7, R^s_{A,6} \Big) & \text{for $i=7, j=6$} \\
        0 & \text{otherwise}
    \end{cases}
\end{equation*}
When computing precision, {\LEA} precision is used instead to evaluate 
accommodated sets. 

\subsection{Computing {\BLANC}}

For this metric we need to determine how the coreference and the non-coreference links in the key and response entities compare.
The link space is large, but let us look, for example, at $N_K(3,7)$ and $N_{R}(3,6)$, the sets of non-coreference links between keys $K_3$ and $K_7$, and response entities $R_{A,3}$ and $R_{A,6}$, respectively. 
Starting with the former, we have:\footnote{We shall use the id of the mentions, single or split-antecedent, 
for a more concise representation.}
\begin{equation*}
    N_K (3,7) = \Big\{ (K^s_3, K^s_7), (K^s_3, 15), (6, K^s_7), (6, 15), (7, K^s_7), (7, 15), (10, K^s_7), (10, 15) \Big\} 
\end{equation*}
The non-coreference links between the response entities $R_{A,3}$ and $R_{A,6}$ are:
\begin{align*}
    N_{R}&(3,6) = \Big\{ (R^s_{A,3}, R^s_{A,6}), (R^s_{A,3}, 15), (6, R^s_{A,6}), \\ & (6, 15), (7, R^s_{A,6}), (7, 15), (10, R^s_{A,6}), (10, 15), (12, R^s_{A,6}), (12, 15) \Big\} 
\end{align*}
Let us now consider how matching links are determined in a recall-based evaluation. 
Two links, one from the key, and the other from the response, whose nodes are regular mentions, get full credit if their mentions match. In our example that happens in 3 cases:
\begin{align*}
    \delta \Big( (6,15), (6,15) \Big) &= 1 \\
    \delta \Big( (7,15), (7,15) \Big) &= 1 \\
    \delta \Big( (10,15), (10,15) \Big) &= 1
\end{align*}
Two links, where one of the nodes is a regular mention, and the other 
an accommodated set, 
are given credit if the regular mentions match and the accommodated sets 
are aligned. 
The allocated credit depends on how well the accommodated sets 
evaluate:
\begin{align*}
    \delta \Big( (K^s_3,15), (R^s_{A,3}, 15) \Big) &= \text{BLANC}_{R_N}(K^s_3, R^s_{A,3}) \\
    \delta \Big( (6,K^s_7), (6,R^s_{A,6}) \Big) &= \text{BLANC}_{R_N}(K^s_7, R^s_{A,6}) \\
    \delta \Big( (7,K^s_7), (7,R^s_{A,6}) \Big) &= \text{BLANC}_{R_N}(K^s_7, R^s_{A,6}) \\
    \delta \Big( (10,K^s_7), (10,R^s_{A,6}) \Big) &= \text{BLANC}_{R_N}(K^s_7, R^s_{A,6})
\end{align*}
Two links both of whose  nodes are 
accommodated sets 
receive credit if the 
accommodated sets 
are aligned, and the score depends on how well the response evaluates against the key:
\begin{equation*}
    \delta \Big( (K^s_3, K^s_7), (R^s_{A,3}, R^s_{A,6}) \Big) = \text{BLANC}_{R_N} \Big(K^s_3, R^s_{A,3} \Big) \times 
     \text{BLANC}_{R_N} \Big(K^s_7, R^s_{A,6} \Big)
\end{equation*}
There is no alignment 
for all  other key and response links,
so no credit can be allocated in these cases, and 
$\delta = 0$. 
Finally, notice we used $R_N$ to compare the entities 
included in the accommodated sets,
as the computations above were used to determine
the credit allocated to non-coreference links in a recall-based evaluation.
Computing the credit for the coreference links involves the same steps, but using $R_C$ instead. And when turning to precision, $P_N$ and $P_C$, the precision related metrics from {\BLANC} are used.

\section{The alternatives}
\label{sec:relatedwork}

There is limited previous work on
split-antecedent anaphora resolution and its evaluation.
We are aware of four proposals, two of which advanced by ourselves in previous work. 

\citet{vala-etal-2016-antecedents} and \citet{yu-etal-2020-plural} only evaluate their models on split-antecedent anaphors (gold mentions only),
and 
compute 
precision, recall, and F1 measures 
based on  the links between  split-antecedent anaphors and their  antecedent. 
%
%
Because they only evaluate on gold split-antecedents, 
their evaluation method does not require any form of alignment or worry about both single and multiple antecedents.

\citet{zhou&choi:COLING2018}  propose a method to evaluate split-antecedent resolution 
using the standard \ACRO{conll} scorer. 
This is done by adding the plural mention to each of the clusters for its atomic elements:
for example, they represent the \textit{\{\{John, Mary\}, They\}} entity as two  gold clusters--\{John, They\} and \{Mary, They\}.
This representation however violates the fundamental assumption behind the notion of coreference chain--that all mentions in the chain refer to the same entity.
In order to compare  our approach to evaluation with theirs, in the next Section we test our state-of-the-art coreference resolver interpreting both single and split antecedent anaphora on the same corpus used in their paper. 



Finally, \citet{yu2021together} proposed an extension of the {\LEA} metric that can also score split-antecedent references. 
The first advantage of the current proposal over our earlier proposal is that the extension proposed in this paper covers \emph{all}  existing metrics for coreference, thus does not require  changing the current 
evaluation paradigm--
the same metrics 
score 
both single and split-antecedent entities. 
Secondly,  \citep{yu2021together} 
 compare  split-antecedents not by  evaluating the entities they refer to 
as done here, but merely the mentions of these entities. In terms of the notation introduced in Section \ref{sec:notation},
the difference is as follows.
Let $K^{s_m}_i$ be the set of mentions of the plural entity $K_i^s$, i.e., the mentions that are explicitly annotated for specifying $K_i^s$ in the text.
The first difference between their approach and ours lies in the alignment step: a gold split-antecedent $K^s_i$ has a matching response split-antecedent $R^s_j$, i.e., $\tau(K^s_i)=R^s_j$, if this is the split-antecedent in the system entities that has the largest number of mentions in common with $K^s_i$, i.e., $R^{s_m}_j = \argmax_{R^{s_m}_{j*}} |K^{s_m}_i \cap R^{s_m}_{j*}| $. Following our presentation of {\LEA} in Section \ref{sec:lea}, the aligned-split antecedents, when evaluating 
recall, 
are assessed based on the recall of their mentions:
\begin{equation*}
    \delta_{i,j} = \frac{|K^{s_m}_i \cap R^{s_m}_{j}|}{|K^{s_m}_i|} 
\end{equation*}
As with our extension, no credit is given for those split-antecedents which are not aligned.
In addition to the alignment step above, the method by \citeauthor{yu2021together} also needs a pre-alignment step on the regular entities using the {\CEAF} score,  which is not required here. The pre-alignment step is deterministic and impose a hard alignment between key and response clusters, and the later scoring step does \textit{not} take into account how good the alignments are. This is problematic as the hard one-to-one constraint is not robust to align multiple equivalent response clusters. Consider the following example.
Suppose we have the following key entities:

\begin{equation*}
\begin{split}
    K_1 &= \{ \text{John}_1, \text{John}_2, \text{him}_5, \text{he}_2\} \\
    K_2 &= \{ \text{Mary}_2, \text{her}_3, \text{She}_4 \} \\
    K_3 &=  \{K_1, K_2\} \oplus \{\text{they}_7\} 
\end{split}
\end{equation*}

And we have the response:
\begin{equation*}
\begin{split}
    R_1 &= \{ \text{John}_1, \text{him}_5\}\\
    R_2 &= \{ \text{John}_2, \text{he}_2\} \\
    R_3 &= \{ \text{Mary}_2, \text{her}_3, \text{She}_4 \} \\
    R_4 &=  \{R_1, R_3\} \oplus \{\text{they}_7\} \\
    R_4' &=  \{R_2, R_3\} \oplus \{\text{they}_7\} 
\end{split}
\end{equation*}

For $R_4$ we have two system predictions ($R_4$ and $R_4'$), which in theory  should be equivalent since they both get the half of the $K_1$ and all of $K_2$. 
However, because the pre-alignment step requires a hard alignment,
either $R_1$ or $R_2$ will be aligned with $K_1$. 
Suppose $R_1$ is aligned with $K_1$; 
the \citeauthor{yu2021together}'s scorer will then give a score of 100\% to $\{R_1, R_3\}$ but 50\% to $\{R_2, R_3\}$. By contrast, the new scorer proposed in this paper will correctly assign the same scores to both interpretations.

In order to compare our new approach to evaluation to the one proposed by \cite{yu2021together}, in  the next Section we use our metrics to score the same system tested by Yu et al on the same corpus.

\section{Using the Metrics for Scoring Generalized Anaphoric Reference}
\label{sec:evaluation}

\begin{table*}[t]
\centering
\begin{subtable}{\textwidth}
\centering
\resizebox{0.95\textwidth}!{
\begin{tabular}{lcccccccccc}
\toprule
&\multicolumn{3}{c}{\bf MUC}&\multicolumn{3}{c}{\bf B$^3$ }&\multicolumn{3}{c}{\bf CEAF$^E$}&\bf CoNLL\\
&\bf R&\bf P&\bf F1&\bf R&\bf P&\bf F1&\bf R&\bf P&\bf F1&\bf F1\\\midrule
Recent 2 & 76.6 & 77.3 & 76.9 & 78.8 & 76.1 & 77.5 & 78.0 & 74.3 & 76.1 & 76.8 \\
Recent 3 & 76.7 & 77.3 & 77.0 & 78.9 & 76.2 & 77.5 & 78.0 & 74.4 & 76.1 & 76.9 \\
Recent 4 & 76.8 & 77.3 & 77.0 & 79.0 & 76.2 & 77.5 & 78.0 & 74.4 & 76.1 & 76.9 \\
Recent 5 & 76.7 & 77.3 & 77.0 & 79.0 & 76.2 & 77.5 & 78.0 & 74.3 & 76.1 & 76.9 \\
Random & 76.5 & 77.1 & 76.8 & 78.8 & 76.0 & 77.4 & 77.9 & 74.2 & 76.0 & 76.7 \\\midrule
Single Ant & 76.3 & 78.4 & 77.4 & 78.7 & 76.9 & 77.8 & 78.0 & 74.4 & 76.2 & 77.1 \\
Best Model & 77.1 & 77.9 & 77.5 & 79.1 & 76.5 & 77.8 & 78.1 & 74.5 & 76.3 & 77.2 \\
Oracle & 77.6 & 78.4 & 78.0 & 79.5 & 76.8 & 78.1 & 78.3 & 74.6 & 76.4 & 77.5 \\
\bottomrule
\end{tabular}}
\caption{\label{table:conll_scorer} {{\MUC}, {\BCUBED}, {\CEAF}$^E$ and \ACRO{conll} F1.}}
\end{subtable}
\begin{subtable}{\textwidth}
\centering
\resizebox{0.95\textwidth}!{
\begin{tabular}{lcccccccccccc}
\toprule
&\multicolumn{3}{c}{\bf CEAF$^M$}&\multicolumn{3}{c}{\bf BLANC}&\multicolumn{3}{c}{\bf LEA ($\beta$=1) }&\multicolumn{3}{c}{\bf LEA($\beta$=10)}\\
&\bf R&\bf P&\bf F1&\bf R&\bf P&\bf F1&\bf R&\bf P&\bf F1&\bf R&\bf P&\bf F1\\\midrule
Recent 2 & 77.2 & 75.4 & 76.3 & 75.3 & 71.5 & 73.4 & 70.3 & 66.9 & 68.6 & 59.7 & 61.2 & 60.5\\
Recent 3 & 77.3 & 75.4 & 76.3 & 75.3 & 71.5 & 73.4 & 70.4 & 66.9 & 68.6 & 60.3 & 61.3 & 60.8\\
Recent 4 & 77.3 & 75.4 & 76.3 & 75.3 & 71.6 & 73.4 & 70.4 & 66.9 & 68.6 & 60.7 & 61.5 & 61.1\\
Recent 5 & 77.3 & 75.4 & 76.3 & 75.3 & 71.6 & 73.4 & 70.4 & 66.9 & 68.6 & 60.7 & 61.1 & 60.9\\
Random & 77.2 & 75.3 & 76.2 & 75.3 & 71.5 & 73.3 & 70.3 & 66.7 & 68.4 & 59.3 & 59.9 & 59.6\\\midrule
Single Ant & 77.2 & 75.8 & 76.5 & 75.2 & 72.3 & 73.7 & 70.3 & 67.4 & 68.8 & 59.0 & 67.4 & 62.9\\
Best Model & 77.4 & 75.7 & 76.6 & 75.5 & 71.9 & 73.6 & 70.7 & 67.2 & 68.9 & 62.8 & 64.7 & 63.7\\
Oracle & 77.8 & 75.9 & 76.8 & 75.9 & 72.1 & 74.0 & 71.2 & 67.6 & 69.4 & 65.9 & 68.4 & 67.1\\
\bottomrule
\end{tabular}}
\caption{\label{table:other_scorer} {{\CEAF}$^M$, {\BLANC} and {\LEA} with different split-antecedent importance ($\beta$).}}
\end{subtable}
\caption{\label{table:system_predictions} 
Evaluation of \citet{yu2021together}'s model and of the baselines on {\ARRAU} using the new Universal Anaphora scorer.
}
\end{table*}

\begin{table*}[t]
\centering
\resizebox{0.95\textwidth}!{
\begin{tabular}{lccccccc}
\toprule
&\bf MUC&\bf B$^3$&\bf CEAF$^E$&\bf CEAF$^M$ &\bf BLANC &\bf LEA &\bf CoNLL\\\midrule
Recent 2 & 25.6 & 22.2 & 15.9 & 23.6 & 15.8 & 21.9 & 21.2 \\
Recent 3 & 27.1 & 24.2 & 20.8 & 26.3 & 18.2 & 23.5 & 24.0 \\
Recent 4 & 28.0 & 25.2 & 21.0 & 27.5 & 17.5 & 24.4 & 24.7 \\
Recent 5 & 26.6 & 23.6 & 19.0 & 26.0 & 15.9 & 22.9 & 23.1 \\
Random & 19.6 & 15.3 & 7.9 & 16.8 & 11.4 & 14.8 & 14.3 \\\midrule
Best Model & 35.8 & 31.9 & 37.0 & 32.8 & 18.2 & 30.9 & 34.9 \\
Oracle & 70.1 & 63.7 & 62.9 & 68.1 & 68.6 & 61.4 & 65.6 \\
\bottomrule
\end{tabular}}
\caption{\label{table:system_predictions_split_only} 
Split-antecedent F1 scores only for \citeauthor{yu2021together}'s systems  on {\ARRAU} evaluated using our extension of the coreference scorers.
}
\end{table*}

Because no generalizations of the existing coreference metrics to cover split-antecedent anaphoric reference was previously proposed, it is not possible to compare our generalization to others. 
However, it is possible to show that, unlike the existing and partial solutions discussed  in the previous Section, the proposed metrics  can be used to compare anaphoric resolvers in exactly  the same way as done using  the existing coreference metrics.

In this Section, our extended metrics were incorporated in the new Universal Anaphora scorer,
a new scorer for anaphora that can also score split antecedent anaphora resolution (as well as non-referring mentions identification, bridging reference resolution and discourse deixis resolution) \citep{yu-et-al:UA-scorer-22}, and this scorer
was then used to score 
the performance of the state of the art system able to carry out split-antecedent anaphora resolution on real data, and to compare it with simpler baselines, both on the corpus used by \cite{yu2021together} and on the corpus used by \cite{zhou&choi:COLING2018}.
%
%

\subsection{The models being compared}

In order to evaluate the generalized coreference metrics on a 
real system output, 
we obtained the best-performing  output and all the baselines from the system by \citet{yu2021together},
the only modern system that can process both single- and split-antecedent anaphors,
and ran our scorer on all six predictions. 
\citeauthor{yu2021together}'s model
is an extension of 
the system 
proposed by \citet{yu-etal-2020-cluster} 
which further interprets 
split-antecedents. 
The model shares most of the network architecture with  \citep{yu-etal-2020-cluster}, but in addition, 
it includes  a dedicated feed-forward network for  split-antecedents. 
The baselines are based on heuristic rules or random selection. 
The same candidate split-antecedent anaphors/singular clusters are used in the baselines as in the best model from \cite{yu2021together}.
Then these baselines attempt to interpret 
the mentions belonging to a small list of plural pronoun which could be interpreted as split antecedent anaphor (e.g., \LINGEX{they}, \LINGEX{their}, \LINGEX{them}, \LINGEX{we})  but were classified as discourse-new by the single-antecedent anaphoric resolver, and attempt to resolve them as split-antecedent anaphors. 
The \NEWTERM{random} baseline randomly assigns two to five antecedents  to these candidate split antecedent anaphors.
After that, the \NEWTERM{recent-x} baseline
uses the x closest singular clusters as antecedents to each chosen anaphor. 
In 
addition, 
we also include scores for the system only resolving single antecedent anaphors (Single Ant) and system augmented with the gold split-antecedent anaphors (Oracle)\footnote{For oracle setting we allow the system to use the gold split-antecedent anaphors annotations when possible. Please note the system still constrained by the quality of singular clusters. This simulates a better system on resolving the split-antecedent anaphors.}.


\subsection{Evaluation on {\ARRAU}}

Table \ref{table:system_predictions} shows the overall scores (single and split-antecedent anaphors) 
 on {\ARRAU}
 evaluated using the new scorer.\footnote{
Following \citet{yu2021together}, we only report scores  for documents in which at least one split-antecedent anaphor is annotated.} 
The general direction of the results doesn't change from those reported in \citet{yu2021together} (see below). 
What changes is that whereas in \citet{yu2021together} the system's performance on split antecedents could only be evaluated using a single, \emph{ad-hoc} metric (see, e.g., Tables 3 and 5 in that paper), thanks to the extension proposed in this paper it is now possible to score both single-antecedent and split-antecedent anaphors using the same metrics. 

The results with all the metrics confirm the results obtained by Yu et al with their specialized metric. 
First of all, 
as already observed by \citet{yu2021together}, the difference between the baselines and the best model is small when  single- and split-antecedent anaphors are evaluated together,
because the number of split-antecedents is low (only 0.8\% of the clusters containing split-antecedents).
Secondly, 
the difference  becomes very clear when considering the performance on split antecedents only (see Table \ref{table:system_predictions_split_only}): up to 20 percentage points in \ACRO{conll} score. 
The Oracle setting has again much better split-antecedent F1 scores, and this results a considerable improvements on all the scores when evaluated with singular clusters.
Confirming 
our hypothesis on the need for partial credit on split-antecedents, 
we find that only 16\% of the split-antecedents were fully resolved by the best model.  
The vast majority of the split-antecedents that were partially resolved rely on the partial credit to get a fair assessment.   

\subsection{Evaluation on {\FRIENDS}}

\begin{table*}[t]
\centering
\begin{subtable}{\textwidth}
\centering
\resizebox{0.95\textwidth}!{
\begin{tabular}{lcccccccccc}
\toprule
&\multicolumn{3}{c}{\bf MUC}&\multicolumn{3}{c}{\bf B$^3$ }&\multicolumn{3}{c}{\bf CEAF$^E$}&\bf CoNLL\\
&\bf R&\bf P&\bf F1&\bf R&\bf P&\bf F1&\bf R&\bf P&\bf F1&\bf F1\\\midrule
Recent 2 & 80.0 & 80.4 & 80.2 & 71.0 & 72.0 & 71.5 & 59.1 & 61.7 & 60.4 & 70.7 \\
Recent 3 & 80.0 & 80.6 & 80.3 & 71.1 & 72.1 & 71.6 & 59.3 & 61.6 & 60.4 & 70.8 \\
Recent 4 & 80.1 & 81.0 & 80.6 & 71.1 & 72.5 & 71.8 & 59.9 & 61.9 & 60.9 & 71.1 \\
Recent 5 & 79.6 & 81.3 & 80.4 & 70.6 & 72.9 & 71.7 & 60.2 & 61.8 & 61.0 & 71.1 \\
Random & 79.8 & 79.9 & 79.9 & 71.0 & 71.4 & 71.2 & 60.6 & 61.3 & 61.0 & 70.7 \\\midrule
Single Ant & 77.9 & 85.0 & 81.3 & 69.1 & 77.0 & 72.8 & 64.0 & 63.0 & 63.5 & 72.6 \\
Best Model & 80.7 & 82.2 & 81.4 & 71.7 & 73.5 & 72.6 & 63.6 & 64.8 & 64.2 & 72.7 \\
Oracle & 81.5 & 83.9 & 82.7 & 72.5 & 75.3 & 73.9 & 66.5 & 65.3 & 65.9 & 74.2 \\
\bottomrule
\end{tabular}}
\caption{\label{table:conll_scorer_friends} {{\MUC}, {\BCUBED}, {\CEAF}$^E$ and \ACRO{conll} F1.}}
\end{subtable}
\begin{subtable}{\textwidth}
\centering
\resizebox{0.95\textwidth}!{
\begin{tabular}{lcccccccccccc}
\toprule
&\multicolumn{3}{c}{\bf CEAF$^M$}&\multicolumn{3}{c}{\bf BLANC}&\multicolumn{3}{c}{\bf LEA ($\beta$=1) }&\multicolumn{3}{c}{\bf LEA($\beta$=10)}\\
&\bf R&\bf P&\bf F1&\bf R&\bf P&\bf F1&\bf R&\bf P&\bf F1&\bf R&\bf P&\bf F1\\\midrule
Recent 2 & 70.8 & 71.8 & 71.3 & 73.1 & 75.9 & 74.5 & 60.9 & 62.0 & 61.4 & 43.1 & 42.8 & 42.9\\
Recent 3 & 70.8 & 71.9 & 71.3 & 73.3 & 76.2 & 74.7 & 61.1 & 62.0 & 61.5 & 43.4 & 42.6 & 43.0\\
Recent 4 & 70.8 & 72.2 & 71.5 & 73.3 & 76.5 & 74.9 & 61.3 & 62.3 & 61.8 & 43.5 & 43.8 & 43.7\\
Recent 5 & 70.5 & 72.3 & 71.4 & 73.2 & 76.9 & 75.0 & 60.9 & 62.5 & 61.7 & 40.9 & 43.9 & 42.3\\
Random & 71.1 & 71.5 & 71.3 & 73.5 & 75.7 & 74.6 & 60.8 & 61.1 & 61.0 & 42.5 & 39.8 & 41.1\\\midrule
Single Ant & 70.4 & 74.7 & 72.5 & 72.4 & 80.1 & 76.0 & 60.5 & 65.5 & 62.9 & 34.8 & 65.5 & 45.5\\
Best Model & 72.0 & 73.3 & 72.6 & 73.6 & 76.8 & 75.2 & 63.0 & 64.2 & 63.6 & 47.6 & 50.5 & 49.0\\
Oracle & 73.0 & 74.1 & 73.6 & 74.6 & 78.1 & 76.3 & 64.6 & 66.2 & 65.4 & 53.4 & 61.2 & 57.1\\
\bottomrule
\end{tabular}}
\caption{\label{table:other_scorer_friends} {{\CEAF}$^M$, {\BLANC} and {\LEA} with different split-antecedent importance ($\beta$).}}
\end{subtable}
\caption{\label{table:system_predictions_friends} 
Evaluation of \citet{yu2021together}'s model and of the baselines on the Friends Corpus using the new Universal Anaphora scorer.
}
\end{table*}

\begin{table*}[t]
\centering
\resizebox{0.95\textwidth}!{
\begin{tabular}{lccccccc}
\toprule
&\bf MUC&\bf B$^3$&\bf CEAF$^E$&\bf CEAF$^M$ &\bf BLANC &\bf LEA &\bf CoNLL\\\midrule
Recent 2 & 43.4 & 37.5 & 30.9 & 42.7 & 28.3 & 36.1 & 37.3 \\
Recent 3 & 45.1 & 38.9 & 32.0 & 43.8 & 30.7 & 37.3 & 38.6 \\
Recent 4 & 44.0 & 37.7 & 31.2 & 42.7 & 30.2 & 36.0 & 37.6 \\
Recent 5 & 42.8 & 36.3 & 28.5 & 41.4 & 30.6 & 34.5 & 35.9 \\
Random & 43.5 & 36.9 & 34.2 & 42.2 & 31.1 & 35.3 & 38.2 \\\midrule
Best Model & 52.3 & 44.9 & 43.3 & 50.8 & 37.8 & 43.3 & 46.8 \\
Oracle & 65.5 & 57.1 & 65.6 & 65.1 & 50.9 & 54.8 & 62.7 \\
\bottomrule
\end{tabular}}
\caption{\label{table:system_predictions_split_only_friends} 
Split-antecedent F1 scores only for \citeauthor{yu2021together}'s systems evaluated on the {\FRIENDS} corpus using our generalization of the coreference scorers.
}
\end{table*}

In order to compare our metrics in practice with the evaluation approach  proposed by \citet{zhou&choi:COLING2018},
we further tested our extended metrics  on the {\FRIENDS} corpus used in \citep{zhou&choi:COLING2018}, 
which contains a larger percentage of split-antecedent anaphors. 
We follow 
\citet{zhou&choi:COLING2018} in using episodes 1 - 19 for training, 20, 21 for development and 22, 23 for testing. The original corpus is annotated for  entity-linking, so the coreference clusters are created by grouping the mentions that refer to the same entity (character in the show). 
14.5\%  of those  clusters contain split-antecedents.
In the original annotation, split-antecedent anaphors represent 9\% of all  mentions; however, 
this is because in the original version of the {\FRIENDS} corpus all subsequent mentions of a set accommodated using a split-antecedent anaphor are also marked as split-antecedents instead of being coreferent with the first mention.
(E.g. in our illustrative example, \LINGEX{they$_7$} and \LINGEX{The two$_{10}$} are treated as single-antecedent by linking them to their$_6$, whereas in the {\FRIENDS} corpus would be annotated as split-antecedent as well.) 
We transformed all of these cases into single-antecedent anaphors; 
after this transformation,  4.1\% of the mentions remain split-antecedent anaphors. 

To obtain the system predictions, we trained the \citet{yu2021together} system on the {\FRIENDS} corpus and computed 
all the baselines in the same way as 
for the {\ARRAU} corpus\footnote{We contacted the authors of \cite{zhou&choi:COLING2018} to obtain their system's outputs but did not get a reply.}. 
As shown in Table \ref{table:system_predictions_friends}, the best model outperforms the baselines by a large margin according to all the metrics 
even though 
the performance 
improvements 
on  split-antecedent anaphors (see Table \ref{table:system_predictions_split_only_friends}) are smaller
than those we observed 
with the {\ARRAU} evaluation. 
This was expected, as the {\FRIENDS} corpus contains many more split-antecedent anaphors. When comparing the best model with the single-antecedent only system (Single Ant), the best model has a better recall but a lower precision, overall having similar F1 scores for most of the matrices. 
This is 
because 
system performance on the split-antecedent part is not good enough to make a clear difference. 
With the oracle setting, however ,
the better performance on split-antecedent anaphors contributed 
to a robust improvement on 
overall performance on both single- and split-antecedent anaphors. This indicates a better split-antecedent anaphora resolver is needed to achieve a significant improvement when compared with systems that only resolve single-antecedent anaphors. If split-antecedent anaphors are the main focus of the evaluation, one can use the split-antecedent F1 scores or the LEA metric with appropriate split-antecedent importance (e.g. $\beta=10$) to prioritise  split-antecedent anaphors.


\section{Scoring other types of split-antecedent anaphora and of anaphora involving accommodation}
\label{sec:scoring-accommodation}

As discussed in the Introduction and in Section \ref{sec:anaphora}, split-antecedent plurals are just one example of anaphoric reference referring to an entity which wasn't previously mentioned, thus requiring accommodation of a new antecedent \citep{beaver&zeevat:accommodation,van-der-sandt:92}.
In all of these cases, the new entity is composed of a part constructed out of the pre-existing discourse model, together with a `coreference chain' part--i.e., the structure proposed here for split antecedent plurals:

\begin{equation*}
    K_i = K^o_i \oplus K^m_{i}\\
\end{equation*}
The difference is the relation linking the new entity to existing  entities in the context. 
For split antecedents, this relation is set membership: the new set is the set of the entities mentioned by the split antecedents. 
This type of accommodation is required not just for plurals, but for discourse deixis as well. 
In the case of bridging references, the relation is associative, not coreference.
In the case of context change accommodation, the new entity is typically the result of an action carried out over the entities in the context.
So, the proposed notation could potentially also serve as the basis for extensions covering these cases. 


\paragraph{Split-antecedent discourse deixis} Another example mentioned in Section \ref{sec:anaphora} of anaphoric reference possibly involving split-antecedents is discourse deixis \CITE{webber:91,kolhatkar-et-al:CL18}, illustrated by \SREF{ex:dd:trains}. 
The Universal Anaphora scorer evaluates discourse deixis exactly in the same way as other types of coreference, the only difference being that the first mention(s) in the chair are utterances rather than nominal phrases. 
The extension proposed in this paper can then be immediately used for the cases of split-antecedent discourse deixis such as \SREF{ex:dd:trains}.

\paragraph{Bridging references}
In the case of split-antecedent anaphora, 
what is accommodated is  a new set composed of already introduced entities. 
In the cases of bridging references, as in \SREF{ex:bridging:2}, we use the notation \LINGEX{[The door]$_j^\text{poss(i)}$} to indicate that the door is a new object, but is related to entity $i$, the house, by a part-of relation.

\begin{EXAMPLE}
\ENEW{ex:bridging:2} John walked towards [the house]$_i$. [The door]$_j^\text{poss(i)}$ was open.
\end{EXAMPLE}
It's not clear however whether such an extension is needed for bridging references, as the evaluation metric proposed by \cite{hou-et-al:CL18} and used both in the 2018 {\CRAC} Shared Task and in the 2021 {\CODICRAC} shared task (and implemented in the {\UA} scorer) \citep{codi-crac-shared-task} appears adequate. 

\paragraph{Context change accomodation} 
A more complex case of anaphoric reference involving accommodation are the cases of \NEWTERM{context change accommodation} discussed by \citet{webber&baldwin:ACL92}, where a new entity, the dough, is obtained by mixing together flour and water.

\begin{EXAMPLE}
\ENEW{ex:context-change} Add  [the water]$_i$ to [the flour]$_j$ little by little.\\
Then work [the dough]$_k^{i+j}$
\end{EXAMPLE}

As far as we are aware context change accommodation  is not annotated in any existing dataset and there is no evaluation method, so developing an approach based on the proposal in this paper would appear to fill a gap in the literature, but such an approach cannot at present be tested.

\section{Conclusion}
\label{sec:conclusion}

In order to push forward the state of the art in anaphora resolution beyond the simplest form of identity anaphora it is not sufficient to create suitable datasets annotated with the more general cases of anaphora, although that is an important effort. 
It is also necessary to develop methods for evaluating the performance of anaphoric resolvers on these cases. 
In this paper we proposed a method for evaluating one of these more general cases--the case of anaphoric 
reference to entities that need introducing in a discourse model via accomodation, exemplified by split-antecedent anaphors but including other cases as well, such as discourse deixis--that is a straightforward extension of 
existing proposals for 
coreference evaluation and thus does not require introducing additional metrics, 
an issue 
in a field already 
over-rich 
with proposals in this direction.

\section*{Acknowledgments}
This research was supported in part by the DALI project, ERC Grant 695662.

\bibliography{dnd}
\end{document}